\documentclass[10pt,journal,compsoc]{IEEEtran}
\ifCLASSOPTIONcompsoc
  \usepackage[noadjust]{cite}
\else
  \usepackage{cite}
\fi
\ifCLASSINFOpdf
   \usepackage[pdftex]{graphicx}
   \graphicspath{{img/}}
\else
\fi
\usepackage{amsmath,amsfonts}
\usepackage{amsthm}
\usepackage{algorithmic}
\usepackage{array}
\usepackage[caption=false,font=normalsize,labelfont=sf,textfont=sf]{subfig}
\usepackage{textcomp}
\usepackage{stfloats}
\usepackage{url}
\usepackage{verbatim}
\usepackage{graphicx}
\def\BibTeX{{\rm B\kern-.05em{\sc i\kern-.025em b}\kern-.08em
    T\kern-.1667em\lower.7ex\hbox{E}\kern-.125emX}}
\usepackage{mydef}

\begin{document}
\title{A Survey on Self-Supervised Graph Foundation Models: Knowledge-Based Perspective}
\author{Ziwen~Zhao$^\dag$,
        Yixin~Su$^\dag$,
        Yuhua~Li\textsuperscript{\Letter},
        Yixiong~Zou,
        Ruixuan~Li,
        and~Rui~Zhang\textsuperscript{\Letter}%
\IEEEcompsocitemizethanks{
    \IEEEcompsocthanksitem Z.~Zhao, Y.~Su, Y.~Li, Y.~Zou, R.~Li, and R.~Zhang are with School of Computer Science and Technology, Huazhong University of Science and Technology. E-mail: \{zwzhao, idcliyuhua, yixiongz, rxli\}@hust.edu.cn, yixin.su@outlook.com, rayteam@yeah.net (\href{http://www.ruizhang.info}{www.ruizhang.info}).
}
\thanks{$^\dag$ Z.~Zhao and Y.~Su are co-first authors.}
\thanks{\textsuperscript{\Letter} Y.~Li and R.~Zhang are corresponding authors.}
\thanks{
This work is supported by the National Key Research and Development Program of China under grant 2024YFC3307900; the National Natural Science Foundation of China under grants 62436003, 62376103, 62206102 and 62302184; the Science and Technology Support Program of Hubei Province under grant 2022BAA046; Hubei science and technology talent service project under grant 2024DJC078; Ant Group through CCF-Ant Research Fund; and the HPC Platform of Huazhong University of Science and Technology.
}
}

\markboth{Journal of \LaTeX\ Class Files,~Vol.~18, No.~9, September~2020}%
{How to Use the IEEEtran \LaTeX \ Templates}

\newtheorem{example}{Example}
\newtheorem{theorem}{Theorem}
\theoremstyle{definition}
\newtheorem{defn}{Definition}

\maketitle

\begin{abstract}
The field of graph foundation models (GFMs) has seen a dramatic rise in interest in recent years. 
Their powerful generalization ability is believed to be endowed by self-supervised pre-training and downstream tuning techniques.
There is a wide variety of knowledge patterns embedded in the graph data, such as node properties and clusters, which are crucial for learning generalized representations for GFMs. 
We present a comprehensive survey of self-supervised GFMs from a novel knowledge-based perspective. 
Our main contribution is a knowledge-based taxonomy that categorizes self-supervised graph models by the specific graph knowledge utilized: microscopic (nodes, links, etc.), mesoscopic (context, clusters, etc.), and macroscopic (global structure, manifolds, etc.). 
It covers a total of 9 knowledge categories and 300 references for self-supervised pre-training as well as various downstream tuning strategies.
Such a knowledge-based taxonomy allows us to more clearly re-examine potential GFM architectures, including large language models (LLMs), as well as provide deeper insights for constructing future GFMs.
\end{abstract}

\begin{IEEEkeywords}
Graph foundation models, self-supervised learning, pre-training, graph neural networks, large language models
\end{IEEEkeywords}

\section{Introduction}

\IEEEPARstart{G}{raphs} are prevalent in various real-world applications. They exhibit diverse knowledge patterns due to the inherent topology~\cite{GNN,DLGsurvey,TKDE22survey}. Moreover, the availability of features and properties associated with nodes and links, such as textual attributes and centrality measures, further enriches the knowledge present in graphs. Over time, deep graph mining techniques have evolved from graph neural networks (GNNs)~\cite{GCN,GAT,GraphSAGE} to graph Transformers (GTs)~\cite{GROVER,GraphFormers} and more recent large language model (LLM)-based graph language models~\cite{GIANT,TAPE,InstructGLM}. They are motivated by capturing more comprehensive knowledge patterns within the graph data, from local relationships to the global structure.

\begin{figure}[]
  \centering
  \includegraphics[scale=0.56]{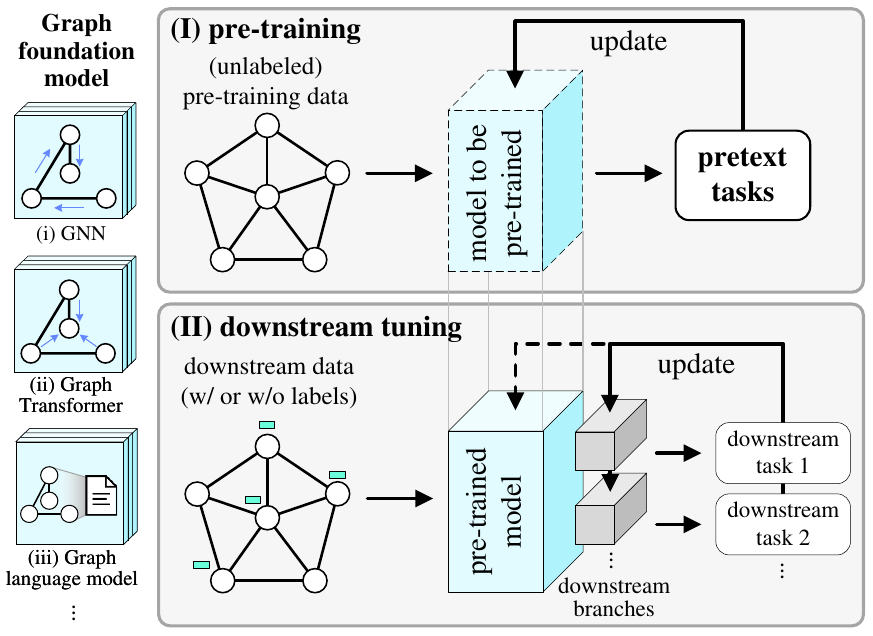}
  \caption{
  How self-supervised GFMs are believed to work: pre-training and downstream tuning. 
  Updating the pre-trained model during downstream tuning is optional depending on the tuning strategy.
  }
  \vspace{-5pt}
  \label{pretext}
\end{figure}

However, when confronted with various downstream task requirements, researchers often encounter graph data that lacks available labels, such as the field of an article in citation networks. 
Fortunately, {\it self-supervised learning} on graphs has emerged as a powerful approach to uncovering underlying patterns in enormous unannotated data~\cite{GAE/VGAE,DGI}. 
SSL methods design unsupervised tasks -- {\it pretext tasks} -- to pre-train a graph model, and adapt the pre-trained model to the specific application scenarios by downstream tuning approaches, as depicted in Fig.~\ref{pretext}. 
Researchers have observed powerful generalization ability within graph models pre-trained with self-supervision~\cite{GFMsurvey}, as they aim to mine the underlying knowledge patterns of graph data instead of solely relying on manual labels that are limited to specific task spaces.
Therefore, self-supervised pre-training and downstream tuning are believed to be the most promising techniques to achieve a {\it graph foundation model} (GFM) -- a highly generalized model that can handle a wide range of application tasks~\cite{FM}.

{\bf Previous efforts}.
The popularity of self-supervised learning and LLMs on graphs in recent years has given rise to a flood of surveys. 
Early efforts~\cite{SSLsurvey2202,TPAMI22survey,TKDE22survey} focus on summarizing general self-supervised graph models. \cite{LGMsurvey,GLM-TKDE24,LLMsurvey2404,LLMsurveyKDD} systematically summarize the trending direction of graphs meet LLMs, shortly after the sensational debut of ChatGPT. The success of LLMs has also activated heated discussions towards GFMs~\cite{GFMsurvey,GFMposition},  summarizing key techniques and principles of learning generalized graph models and providing outlooks towards the realization of GFMs. 
Despite the promising work, we reveal three major shortcomings of the existing surveys:

(1) {\it Lack of comprehensiveness}:
existing surveys in the field of self-supervised graph learning~\cite{SSLsurvey2202,TPAMI22survey,TKDE22survey} do not cover the latest progress in this fast-developing field. For example, none of these surveys have discussed the new achievements of masked graph autoencoders~\cite{GraphMAE} and learning graph manifolds~\cite{DSGC}. 
A recent survey~\cite{GCLsurvey} includes cutting-edge developments in graph contrastive learning, yet it focuses on real-world applications rather than realizing GFMs.

(2) {\it Unclear categorization}:
existing surveys~\cite{GFMsurvey,TPAMI22survey,TKDE22survey} broadly categorize graph pre-training methods as ``generative -- contrastive (predictive)''. 
This rough categorization is insufficient to capture the unique characteristics of graphs, which have diverse knowledge patterns embedded in their structure and properties. For instance, predicting {\it links} requires local relationships between nodes, whereas predicting {\it clusters} requires the node distribution on the entire graph. However, both generative and contrastive (predictive) frameworks can utilize the knowledge of links~\cite{GAE/VGAE,GraphFormers} and clusters~\cite{M3S,CommDGI}, which the aforementioned taxonomy fails to distinguish. 
On the other hand, recent surveys of GFMs only give a brief summary of existing pre-training and downstream tuning methods: \cite{GFMsurvey} puts the emphasis on the architecture design of graph models, while \cite{LGMsurvey,GFMposition} are closer to outlooks towards future directions of GFMs. 

(3) {\it Limited to specific architectures}:
the aforementioned graph self-supervised learning surveys are limited to GNNs/GTs only. On the other hand, LLM-based surveys~\cite{GLM-TKDE24,LLMsurvey2404,LLMsurveyKDD} overemphasize the language model architectures and textual attributes of graphs while overlooking other structural patterns. A recent GFM survey~\cite{GFMsurvey} categorizes existing studies into three groups of GNN, LLM, and GNN+LLM, still limited by specific backbone architectures instead of an in-depth perspective towards the ultimate goal -- mining generalized graph knowledge. 
As language models are not designed for mining various types of graph knowledge, it still remains an unanswered question if LLMs are ideal architectures for GFMs. If other promising generalized architectures showed up in the near future (which is happening right now), their architecture-based taxonomy might no longer apply.

{\bf Our contributions}. 
Considering the aforementioned issues, it is necessary to provide a comprehensive survey of self-supervised graph models with a clearer categorization and taxonomy, which will offer a better understanding and greater insight into how future GFMs work. 
We first propose a {\it knowledge-based taxonomy} that categorizes self-supervised graph pre-training based on the types of knowledge utilized: 
{\it microscopic} pre-training (Section \ref{3}) that focuses on individual nodes and links; 
{\it mesoscopic} pre-training (Section \ref{4}) that focuses on local relationships in the graph, such as context and clusters; 
and {\it macroscopic} pre-training (Section \ref{5}) that focuses on the structure and the manifold underlying the entire graph. 
Such a knowledge-based taxonomy provides a unified perspective to analyze the pre-training and downstream tuning strategies (Section \ref{6}) of both GNNs/GTs and recent graph language models (Section \ref{7}), providing valuable insights for the future directions of GFMs (Section \ref{8}). 
Our knowledge-based perspective is also architecture-agnostic, compared to existing surveys which are applicable to only certain types of architectures. 
Therefore, we provide a more systematic view covering a much wider range of graph models. 
As illustrated in Fig.~\ref{pttaxonomy}, we analyze 9 knowledge categories and 300 references ranging from the 2010s to 2025, which are to our knowledge the most detailed categorization of self-supervised GFMs. 
All papers included are summarized as tables in Appendix~\ref{A} for better comparison. 
We also collate more than 500 relevant papers and list them on GitHub\footnote{\url{https://github.com/Newiz430/Pretext}}. 
We hope this survey will help researchers exploit more powerful GFMs by exploiting graph-specific knowledge.

\tikzstyle{nodefeatureleaf}=[draw=nodefeatureborderclr,
    rounded corners,minimum height=1em,
    fill=nodefeatureclr!40,text opacity=1, align=center,text=black,align=left,font=\scriptsize,
    inner xsep=3pt,
    inner ysep=1pt,
]
\tikzstyle{nodepropertyleaf}=[draw=nodepropertyborderclr,
    rounded corners,minimum height=1em,
    fill=nodepropertyclr!40,text opacity=1, align=center,text=black,align=left,font=\scriptsize,
    inner xsep=3pt,
    inner ysep=1pt,
]
\tikzstyle{linkleaf}=[draw=linkborderclr,
    rounded corners,minimum height=1em,
    fill=linkclr!40,text opacity=1, align=center,text=black,align=left,font=\scriptsize,
    inner xsep=3pt,
    inner ysep=1pt,
]
\tikzstyle{contextleaf}=[draw=contextborderclr,
    rounded corners,minimum height=1em,
    fill=contextclr!40,text opacity=1, align=center,text=black,align=left,font=\scriptsize,
    inner xsep=3pt,
    inner ysep=1pt,
]
\tikzstyle{longrangesimilarityleaf}=[draw=longrangesimilarityborderclr,
    rounded corners,minimum height=1em,
    fill=longrangesimilarityclr!40,text opacity=1, align=center,text=black,align=left,font=\scriptsize,
    inner xsep=3pt,
    inner ysep=1pt,
]
\tikzstyle{motifleaf}=[draw=motifborderclr,
    rounded corners,minimum height=1em,
    fill=motifclr!40,text opacity=1, align=center,text=black,align=left,font=\scriptsize,
    inner xsep=3pt,
    inner ysep=1pt,
]
\tikzstyle{clusterleaf}=[draw=clusterborderclr,
    rounded corners,minimum height=1em,
    fill=clusterclr!40,text opacity=1, align=center,text=black,align=left,font=\scriptsize,
    inner xsep=3pt,
    inner ysep=1pt,
]
\tikzstyle{globalstructureleaf}=[draw=globalstructureborderclr,
    rounded corners,minimum height=1em,
    fill=globalstructureclr!40,text opacity=1, align=center,text=black,align=left,font=\scriptsize,
    inner xsep=3pt,
    inner ysep=1pt,
]
\tikzstyle{manifoldleaf}=[draw=manifoldborderclr,
    rounded corners,minimum height=1em,
    fill=manifoldclr!40,text opacity=1, align=center,text=black,align=left,font=\scriptsize,
    inner xsep=3pt,
    inner ysep=1pt,
]
\tikzstyle{microscopicmiddle}=[draw=microscopicclr,
    rounded corners,minimum height=1em,
    text opacity=1,fill=nodepropertyclr!20,align=center,text=black,align=left,font=\scriptsize,
    inner xsep=3pt,
    inner ysep=1pt,
]
\tikzstyle{mesoscopicmiddle}=[draw=mesoscopicclr,
    rounded corners,minimum height=1em,
    text opacity=1,fill=longrangesimilarityclr!20,align=center,text=black,align=left,font=\scriptsize,
    inner xsep=3pt,
    inner ysep=1pt,
]
\tikzstyle{macroscopicmiddle}=[draw=macroscopicclr,
    rounded corners,minimum height=1em,
    text opacity=1, fill=manifoldclr!20,align=center,text=black,align=left,font=\scriptsize,
    inner xsep=3pt,
    inner ysep=1pt,
]
\tikzstyle{root}=[draw=black,
    rounded corners,minimum height=1em,
    fill=output-white!40,text opacity=1, align=center,
    fill opacity=.5,  text=black,align=left,font=\scriptsize,
    inner xsep=3pt,
    inner ysep=1pt,
]

\begin{figure*}[ht]
\centering
\begin{tikzpicture}
\node[anchor=north west] (tree) {
\begin{forest}
  for tree={
  forked edges,
  grow=east,
  reversed=true,
  anchor=base west,
  parent anchor=east,
  child anchor=west,
  base=middle,
  font=\scriptsize,
  rectangle,
  line width=0.7pt,
  draw=black,
  rounded corners,
  align=left,
  minimum width=2em,
  s sep=3pt,
  l sep=9pt,
  inner xsep=3pt,
  inner ysep=1pt,
  },
  where level=1{text width=4.5em}{},
  where level=2{text width=6em,font=\scriptsize}{},
  where level=3{font=\scriptsize}{},
  where level=4{font=\scriptsize}{},
  where level=5{font=\scriptsize}{},
  [Self-supervised graph pre-training, root, rotate=90, anchor=north, edge=black
    [Microscopic tasks\\(Section~\ref{3}), microscopicmiddle, edge=black,text width=6em
        [Node features\\(Section~\ref{3.1}), nodefeatureleaf, text width=5.3em, edge=black!20!nodefeatureborderclr
            [Feature prediction, nodefeatureleaf, text width=6em, edge=black!20!nodefeatureborderclr
                [MGAE~\cite{MGAE}{,} GALA~\cite{GALA}{,} Graph-Bert~\cite{Graph-Bert}{,} GMI~\cite{GMI}{,} AttrMask~\cite{GNNpretrain}{,} GraphComp~\cite{SS-GCN}{,} \\AttributeMask~\cite{SelfTask}{,} LaGraph~\cite{LaGraph}{,} SLAPS~\cite{SLAPS}{,} GPT-GNN~\cite{GPT-GNN}{,} GraphMAE~\cite{GraphMAE}{,} \\GraphMAE2~\cite{GraphMAE2}{,} HGMAE~\cite{HGMAE}{,} Mole-BERT~\cite{Mole-BERT}{,} DiscoGNN~\cite{DiscoGNN}, nodefeatureleaf, text width=28.2em, edge=black!20!nodefeatureborderclr
                ]
            ]
            [Node instance discrimination, nodefeatureleaf, text width=9.7em, edge=black!20!nodefeatureborderclr
                [GRACE~\cite{GRACE}{,} GCA~\cite{GCA}{,} ProGCL~\cite{ProGCL}{,} MERIT~\cite{MERIT}{,} COSTA~\cite{COSTA}{,} CGI~\cite{CGI}{,} \\BGRL~\cite{BGRL}{,} SGRL~\cite{SGRL}{,} SUGRL~\cite{SUGRL}{,} SimGCL~\cite{SimGCL}{,} LightGCL~\cite{LightGCL}{,} \\ImGCL~\cite{ImGCL}{,}  GRADE~\cite{GRADE}{,} SP-GCL~\cite{SP-GCL}{,} HASH-CODE~\cite{HASH-CODE}, nodefeatureleaf, text width=24.5em, edge=black!20!nodefeatureborderclr
                ]
            ]
            [Dimension discrimination, nodefeatureleaf, text width=8.5em, edge=black!20!nodefeatureborderclr
                [G-BT~\cite{G-BT}{,} CCA-SSG~\cite{CCA-SSG}{,} VICReg~\cite{VICReg}{,} LogDet~\cite{LogDet} , nodefeatureleaf, text width=25.7em, edge=black!20!nodefeatureborderclr
                ]
            ]
        ]
        [Node properties\\(Section~\ref{3.2}), nodepropertyleaf, text width=5.3em, edge=black!20!nodepropertyborderclr
            [ScoreRank~\cite{UPGCN}{,} NodeProperty~\cite{SelfTask}{,} NWR-GAE~\cite{NWR-GAE}{,} MaskGAE~\cite{MaskGAE}{,} PIGAE~\cite{PIGAE}{,} CenPre~\cite{CenPre}, nodepropertyleaf, text width=35.85em, edge=black!20!nodepropertyborderclr
            ]
        ]
        [Links\\(Section~\ref{3.3}), linkleaf, text width=5.3em, edge=black!20!linkborderclr
            [GAE~\cite{GAE/VGAE}{,} VGAE~\cite{GAE/VGAE}{,} ARGA~\cite{ARGA/ARVGA}{,} ARVGA~\cite{ARGA/ARVGA}{,} SIG-VAE~\cite{SIG-VAE}{,}  EdgeMask~\cite{SelfTask}{,} GPPT~\cite{GPPT}{,} S2GAE~\cite{S2GAE}{,} \\MaskGAE~\cite{MaskGAE}{,} Bandana~\cite{Bandana}{,} 
            CG\textsuperscript{3}~\cite{CG3}{,} SuperGAT~\cite{SuperGAT}{,} PIGAE~\cite{PIGAE}{,} ASD-VAE~\cite{ASD-VAE}{,} D-VGAE~\cite{D-VGAE}, linkleaf, text width=35.8em, edge=black!20!linkborderclr]
        ]
    ]
    [Mesoscopic tasks\\(Section~\ref{4}), mesoscopicmiddle, edge=black, text width=6em
        [Context\\(Section~\ref{4.1}), contextleaf, text width=5.3em, edge=black!20!contextborderclr
                [GraphSAGE~\cite{GraphSAGE}{,} EGI~\cite{EGI}{,} DGSI~\cite{DGSI}{,} COLES~\cite{COLES}{,} GLEN~\cite{GLEN}{,} Self-Pro~\cite{Self-Pro}{,} ContextPred~\cite{GNNpretrain}{,} GCC~\cite{GCC}{,} \\S\textsuperscript{3}-CL~\cite{S3-CL}{,}
                Graph-MLP~\cite{Graph-MLP}{,} N2N~\cite{N2N}{,} Subg-Con~\cite{Subg-Con}{,} AFGRL~\cite{AFGRL}{,} HGRL~\cite{HGRL}{,} BSG~\cite{BSG}, contextleaf, text width=35.8em, edge=black!20!contextborderclr] 
        ]
        [Long-range\\similarities\\(Section~\ref{4.2}), longrangesimilarityleaf, text width=5.3em, edge=black!20!longrangesimilarityborderclr
            [Similarity prediction, longrangesimilarityleaf, text width=6.7em, edge=black!20!longrangesimilarityborderclr
                [S\textsuperscript{2}GRL~\cite{S2GRL}{,} Graph-Bert~\cite{Graph-Bert}{,} PairwiseDistance~\cite{SelfTask}{,} PairwiseAttrSim~\cite{SelfTask}{,} AGE~\cite{AGE}, longrangesimilarityleaf, text width=27.5em, edge=black!20!longrangesimilarityborderclr]
            ]
            [Similarity graph alignment, longrangesimilarityleaf, text width=8.9em, edge=black!20!longrangesimilarityborderclr
                [AM-GCN~\cite{AM-GCN}{,} DLR-GAE~\cite{DLR-GAE}{,} ASP~\cite{ASP}{,} MVMI-FT~\cite{MVMI-FT}{,} AEGCL~\cite{AEGCL}, longrangesimilarityleaf, text width=25.3em, edge=black!20!longrangesimilarityborderclr]
            ]
        ]
        [Motifs\\(Section~\ref{4.3}), motifleaf, text width=5.3em, edge=black!20!motifborderclr
            [GROVER~\cite{GROVER}{,} MGSSL~\cite{MGSSL}{,} GraphFP~\cite{GraphFP}{,} MoAMa~\cite{MoAMa}{,} DGPM~\cite{DGPM}{,} MotifRGC~\cite{MotifRGC}{,} MICRO-Graph~\cite{MICRO-Graph}{,} \\CTAug~\cite{CTAug}, motifleaf, text width=35.85em, edge=black!20!motifborderclr]
        ]
        [Clusters\\(Section~\ref{4.4}), clusterleaf, text width=5.3em, edge=black!20!clusterborderclr
            [Node clustering, clusterleaf, text width=5.3em, edge=black!20!clusterborderclr
                [M3S~\cite{M3S}{,} NodeCluster~\cite{SS-GCN}{,} GraphLoG~\cite{GraphLoG}{,} HomoGCL~\cite{HomoGCL}{,} MGSE~\cite{MGSE}{,} \\CARL-G~\cite{CARL-G}{,} CommDGI~\cite{CommDGI}{,} S$^3$-CL~\cite{S3-CL}{,} DCGL~\cite{DCGL}, clusterleaf, text width=28.9em, edge=black!20!clusterborderclr]
            ]
            [Graph partitioning, clusterleaf, text width=6.2em, edge=black!20!clusterborderclr
                [ClusterDetect~\cite{UPGCN}{,} GraphPar~\cite{SS-GCN}{,} Distance2Clusters~\cite{SelfTask}{,} DGVAE~\cite{DGVAE}{,} \\Mask-GVAE~\cite{Mask-GVAE}{,} gCooL~\cite{gCooL}{,} CSGCL~\cite{CSGCL}{,} StructComp~\cite{StructComp}, clusterleaf, text width=28em, edge=black!20!clusterborderclr]
            ]
        ]
    ]
    [Macroscopic tasks\\(Section~\ref{5}), macroscopicmiddle, edge=black, text width=6em
        [Global structure\\(Section~\ref{5.1}), globalstructureleaf, text width=5.3em, edge=black!20!globalstructureborderclr
            [Graph instance discrimination, globalstructureleaf, text width=10em, edge=black!20!globalstructureborderclr
                [GraphCL~\cite{GraphCL}{,} JOAO~\cite{JOAO}{,} AD-GCL~\cite{AD-GCL}{,} SimGRACE~\cite{SimGRACE}{,}  \\DGI~\cite{DGI}{,} InfoGraph~\cite{InfoGraph}{,} MVGRL~\cite{MVGRL}{,} GRV~\cite{GRV}{,} GGD~\cite{GGD}{,} \\D-SLA~\cite{D-SLA}{,} CGC~\cite{CGC}{,} SPAN~\cite{SPAN}, globalstructureleaf, text width=24.2em, edge=black!20!globalstructureborderclr]
            ]
            [Graph similarity prediction, globalstructureleaf, text width=9em, edge=black!20!globalstructureborderclr
                [KernelPred~\cite{KernelPT}{,} D-SLA~\cite{D-SLA}{,} HTML~\cite{HTML}, globalstructureleaf, text width=25.2em, edge=black!20!globalstructureborderclr]
            ]
        ]
        [Manifolds\\(Section~\ref{5.2}), manifoldleaf, text width=5.3em, edge=black!20!manifoldborderclr
            [HGCL~\cite{HGCL}{,} DSGC~\cite{DSGC}{,} SelfMGNN~\cite{SelfMGNN}{,} Graph-JEPA~\cite{Graph-JEPA}{,} HDM-GAE~\cite{HDM-GAE}{,} RiemannGFM~\cite{RiemannGFM}, manifoldleaf, text width=35.85em, edge=black!20!manifoldborderclr]
        ]
    ]
  ]
\end{forest}
};
\node[anchor=north west, xshift=0cm, yshift=-7pt, opacity=1] at (tree.north west) {
\includegraphics[scale=1]{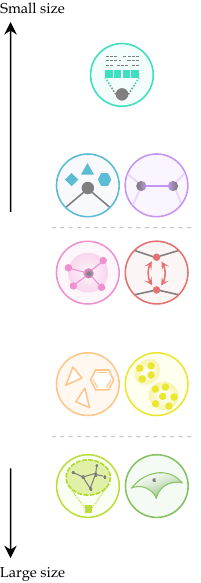}};
\end{tikzpicture}
\caption{
\conditionalblue{
Our knowledge-based taxonomy of self-supervised graph pre-training with representative literature.
} 
}
\label{pttaxonomy}
\end{figure*}

\section{Preliminary}

This section provides basic concepts related to our topic. 

\emph{Graph.}
Graph is a data structure consisting of a node (vertex) set and an edge (link) set $\mathcal{G}=(\mathcal{V}, \mathcal{E})$. The adjacency matrix $\mathbf{A}\in \lbrace  0,1 \rbrace^{n \times n}$ indicates if two nodes are connected by a link. 
For an attributed graph, each node is associated with a row of the feature matrix \smash{$\mathbf{X} \in \mathbb{R}^{n \times d}$}. 
For every node $i \in \mathcal{V}$, its (undirected) neighborhood is \smash{$\mathcal{N}_i = \lbrace j\in\mathcal{V} \vert A_{i,j}=1\rbrace$}. 
A graph dataset can contain one graph only or multiple relatively small graphs. 

\emph{Graph model and graph foundation model (GFM).}
A graph model is an encoding function $\mathbf{Z}=f(\mathcal{G};\Theta)$ that can be parameterized by GNNs~\cite{GCN,GAT,GraphSAGE}, GTs~\cite{GROVER,GraphFormers}, graph language models~\cite{GIANT,TAPE,InstructGLM}, etc.
A GFM is an (ideal) graph model pre-trained on various kinds of unsupervised graph data to handle various types of graph-related tasks~\cite{GFMsurvey}. 

\emph{Pre-training task (pretext).}
A pretext $\mathcal{L} \in \mathcal{T}$ is a self-supervised task performed during the pre-training phase of a graph model, where $\mathcal{T}$ represents the pretext task set. A pretext should meet two conditions: (1) during the self-supervised pre-training, no manual-labeled data is used;
(2) its goal is to achieve improved performance on one or multiple downstream tasks $\check{\mathcal{L}}$:
\begin{equation}\label{gssl}
\resizebox{.91\linewidth}{!}{$\displaystyle
\sum_{\check{\mathcal{L}} \in \check{\mathcal{T}}}{\min_{\Phi,\Theta^*}\check{\mathcal{L}}(\check{f} \cdot f^*, \check{\mathcal{G}},\check{\mathcal{Y}})}, \ s.t. \ f^* = {\sum_{\mathcal{L} \in \mathcal{T}}{\arg \min_{\Theta}\mathcal{L}(f, \mathcal{G})}}
$}
\end{equation}
where \smash{$\check{f}(\check{\mathcal{G}};\Phi)$} denotes some optional downstream branches. 
``\smash{$\Theta^*$}'' is optional depending on whether the pre-trained model parameters are tuned for downstream tasks. 
``\smash{$\check{\mathcal{Y}}$}'' is also optional depending on whether task-specific labels are used, also known as supervised fine-tuning (SFT).

\section{Microscopic Pre-training Tasks} \label{3}

Microscopic pre-training tasks treat nodes or edges as individual instances. They extract features, properties, and local relationships between these instances. 

\subsection{Node Features} \label{3.1}

Node features are a rich source of semantic information in attributed graphs, encoding domain-specific knowledge such as textual content in citation networks or chemical properties in molecular graphs.
The expressiveness and utility of these features heavily depend on their origin and the encoding methods employed.
\\[5pt]
{\bf Feature prediction.} \label{3.1.1}
Feature prediction serves as a fundamental pretext task in graph autoencoding methods like MGAE~\cite{MGAE}, GALA~\cite{GALA}, and Graph-Bert~\cite{Graph-Bert}. These methods reconstruct low-dimensional node representations and match them with the original feature size, minimizing the reconstruction error such as MSE: $\mathcal{L} = \mathbb{E}_{i \in \mathcal{V}}[\Vert \boldsymbol{X}_i - \hat{\boldsymbol{X}}_i\Vert^2]$,
while GMI~\cite{GMI} maximizes the mutual information between the original graph and the output representations by a discriminator network. 

Another kind of prediction task focuses on {\it feature denoising}, where noise is first added to the original features \smash{$\tilde{\mathbf{X}}=\mathbf{X}+\epsilon$}, and then the model is tasked with recovering the original noise-free features. 
The success of masked language/image modeling~\cite{BERT,MAE} has led to the rise of {\it masked feature prediction}, also known as masked autoencoding or graph completion~\cite{SS-GCN}. 
Methods like AttrMask~\cite{GNNpretrain,SelfTask}, LaGraph~\cite{LaGraph}, and SLAPS~\cite{SLAPS} sample a noise matrix from a Bernoulli distribution \smash{$\mathbf{M} \in \lbrace 0,1 \rbrace^{n \times d}$} and obtain masked features $\tilde{\mathbf{X}} = \mathbf{M} \circ \mathbf{X}$. Then, the original features are reconstructed by an MSE loss.
GPT-GNN~\cite{GPT-GNN} adopts an autoregressive masking approach, where the masked node attributes and their corresponding edges are generated one-by-one.
Recent GraphMAE series~\cite{GraphMAE,GraphMAE2} introduces a scaled cosine error with a focusing parameter $\lambda$ to adjust the weight of each sample:
\begin{equation}\label{sce}
\resizebox{.91\linewidth}{!}{$\displaystyle
\mathcal{L}=\mathbb{E}_{q((1-\mathbf{M}) \circ \mathbf{X}|\mathbf{X})} \left[1- \left(\frac{\mathbf{X}^\top f(\mathbf{M} \circ \mathbf{X},\mathbf{A};\Theta)}{\Vert\mathbf{X}\Vert \Vert f(\mathbf{M} \circ \mathbf{X},\mathbf{A};\Theta)\Vert} \right)^\lambda\right]
$}
\end{equation}
and this has inspired various new-generation masked autoencoder architectures~\cite{HGMAE,Mole-BERT,UniGraph}. 
DiscoGNN~\cite{DiscoGNN} first randomly replaces nodes with different ones. Then, it learns to find and reconstruct the replaced nodes.
%
\\[5pt]
\conditionalblue{
{\bf Node instance discrimination.} \label{3.1.3}
Instance discrimination, also referred to as ``contrastive learning'', has achieved significant success in the visual domain~\cite{MoCo,SimCLR} and subsequently becomes a fundamental and general task for graph pre-training. 
Node instance discrimination aims to perform instance discrimination between node pairs. The workflow involves creating two perturbed versions (views) of an original graph $\mathcal{G}^\text{\romannumeral1},\mathcal{G}^\text{\romannumeral2}$. 
Nodes at the same position across views \smash{$(\boldsymbol{Z}^\text{\romannumeral1}_{i},\boldsymbol{Z}^\text{\romannumeral2}_{i})$} form positive pairs, 
while others \smash{$(\boldsymbol{Z}^{\text{\romannumeral1}}_{i},\boldsymbol{Z}^\text{\romannumeral2}_{j})$} or \smash{$(\boldsymbol{Z}^\text{\romannumeral1}_{i},\boldsymbol{Z}^\text{\romannumeral1}_{j})$)} are randomly sampled as negative pairs. 
The goal is to maximize the similarity between positive pairs and minimize it between negative ones, allowing for the learning of general and perturbation-invariant representations.
} 

The simplest form of node-level instance discrimination is to minimize the MSE between positive pairs $\mathcal{L}=\mathbb{E}_{i \in \mathcal{V}}[\Vert {\boldsymbol{Z}^\text{\romannumeral1}_{i}} - \boldsymbol{Z}^\text{\romannumeral2}_{i}\Vert^2]$. 
However, it can suffer from representation degeneration, i.e., the output may degenerate to a constant vector regardless of the input. This is often addressed by combining it with other tasks~\cite{LaGraph,GraphMAE2,CCA-SSG,VICReg}. 
By contrast, {\it mutual information} (MI) provides a more effective criterion by capturing non-linear statistical dependence between node instances~\cite{MINE}:
\begin{equation}\label{mi}
I(\boldsymbol{Z}^\text{\romannumeral1}_{i};\boldsymbol{Z}^\text{\romannumeral2}_{j}) = D_{\text{KL}}[p(\boldsymbol{Z}^\text{\romannumeral1}_{i},\boldsymbol{Z}^\text{\romannumeral2}_{i}) \Vert p(\boldsymbol{Z}^\text{\romannumeral1}_{i})p(\boldsymbol{Z}^\text{\romannumeral2}_{i})]
\end{equation}
Calculating MI in a high-dimension space is a challenging task~\cite{MINE}. Therefore, various techniques have been proposed to estimate MI. These techniques mainly include:

(1) \emph{Jenson-Shannon (JS) estimator}~\cite{f-GAN}: 
it replaces the KL divergence in \eqref{mi} with JS divergence and approximates the distributions usually by a discriminator network $\mathcal{D}$: 
\begin{equation}\label{js}
\resizebox{0.91\linewidth}{!}{$\displaystyle
\mathcal{L}=-\mathbb{E}_{i\in \mathcal{V}}[\sigma_{+}(-\mathcal{D}(\boldsymbol{Z}^\text{\romannumeral1}_{i},\boldsymbol{Z}^\text{\romannumeral2}_{i})) + \mathbb{E}_{j\in \mathcal{V}_i^-}[\sigma_{+}(\mathcal{D}(\boldsymbol{Z}^\text{\romannumeral1}_{i},\boldsymbol{Z}_{\cdot;j}))]]
$}
\end{equation}

(2) \emph{InfoNCE estimator}~\cite{InfoNCE}: it is based on the Noise Contrastive Estimation (NCE) loss. Formally:
\begin{equation}\label{infonce}
\resizebox{0.91\linewidth}{!}{$\displaystyle
\mathcal{L}\!=\!-\mathbb{E}_{i\in \mathcal{V}}\!\left[\log \!\frac{\exp(\langle\boldsymbol{Z}^\text{\romannumeral1}_{i},\!\boldsymbol{Z}^\text{\romannumeral2}_{i}\rangle)}{\sum_{j\ne i}\!{\exp(\langle\boldsymbol{Z}^\text{\romannumeral1}_{i},\!\boldsymbol{Z}^\text{\romannumeral1}_{j}\rangle)}\!+\!\sum_{j=1}^{n}\!{\exp(\langle\boldsymbol{Z}^\text{\romannumeral1}_{i},\!\boldsymbol{Z}^\text{\romannumeral2}_{j}\rangle)}}\right]
$}
\end{equation}
where $\langle\cdot,\cdot\rangle$ is the relative similarity of two samples.
This estimator is well-known in representative models like GRACE~\cite{GRACE}, GCA~\cite{GCA}, ProGCL~\cite{ProGCL}, and more~\cite{MERIT,COSTA,CGI}.

(3) \emph{Triplet (margin) estimator}~\cite{FaceNet}: some contrastive frameworks like SUGRL~\cite{SUGRL} 
employ a triplet loss to contrast between the anchor-positive pairs $(\mathbf{Z},\mathbf{Z}^+)$ the anchor-negative pairs $(\mathbf{Z},\mathbf{Z}^-)$:
\begin{equation}\label{triplet}
\mathcal{L} = \mathbb{E}_{i\in \mathcal{V}}\left[\langle\boldsymbol{Z}_i,\boldsymbol{Z}^{+}_{i}\rangle-\langle\boldsymbol{Z}_i,\boldsymbol{Z}^{-}_{i}\rangle + \epsilon\right]
\end{equation}
where $\epsilon$ denotes the distance margin, controlling the lower bound of distance between positive and negative samples.

There are other instance discrimination objectives that have achieved competitive performance in learning node features. 
For example, the bootstrapping loss~\cite{BYOL,BGRL,SGRL} generally computes the cosine similarity $\mathcal{L} = -\mathbb{E}_{i\in \mathcal{V}}[{\boldsymbol{Z}^\text{\romannumeral1}_{i}{p(\boldsymbol{Z}^\text{\romannumeral2}_{i})}^\top}/{\Vert\boldsymbol{Z}^\text{\romannumeral1}_{i}\Vert\Vert p(\boldsymbol{Z}^\text{\romannumeral2}_{i})\Vert}]$ between two asymmetric and momentum-updated networks, aligned by a projector $p(\cdot)$. 
Other examples include Bayesian Personalized Ranking loss (BPR)~\cite{BPR,SimGCL,LightGCL} and population spectral contrastive loss~\cite{SPCL,SP-GCL,HASH-CODE}. 

\begin{figure}[]
  \centering
\includegraphics[scale=0.84]{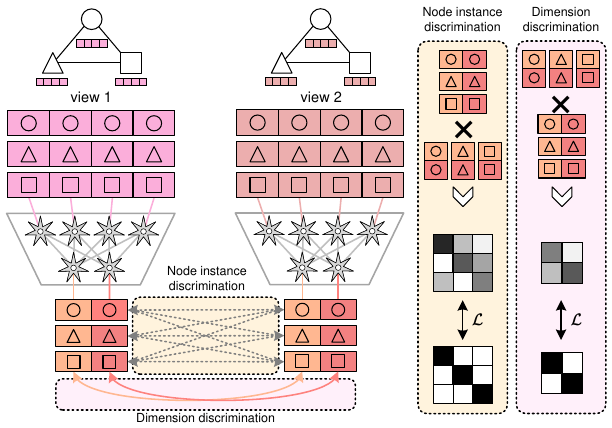}
\caption{ 
An illustration of discrimination tasks between node features. The similarity is defined as the dot-product between embeddings $\mathbf{Z}$. 
Node instance discrimination performs row-wise contrast, which computes the similarity between every pair of node embeddings (distinguished by the node shape: circle/triangle/square).
Dimension discrimination performs column-wise contrast, which computes the similarity between every pair of node dimensions (distinguished by the embedding color: orange/red).
}
\vspace{-5pt}
\label{discrimination}
\end{figure}

Node instance discrimination is one of the most popular and generalizable tasks beyond graph pre-training.
Recent literature focusing on downstream tuning has unified multiple tasks such as link prediction, node classification, and graph classification into node instance discrimination~\cite{GraphPrompt,Self-Pro,IGAP}, as an impactful self-supervised tuning strategy.
\\[5pt]
{\bf Dimension discrimination.} \label{3.1.4}
Dimension discrimination focuses on distinguishing different dimensions of node representations.
This can be seen as a column-wise discrimination approach, as shown in Fig.~\ref{discrimination}. 
Similar to node instance discrimination, it begins with two augmented views. 
Then, it maximizes the correlation between corresponding dimensions and minimizes it between different ones. 
This strategy is initially proposed by Barlow Twins~\cite{BarlowTwins} and then introduced to the graph domain by G-BT~\cite{G-BT}, which considers the similarity between dimensions of different instances. On the other hand, CCA-SSG~\cite{CCA-SSG} 
focuses on the dimensional similarity within a single instance via a covariance regularization term: 
\begin{equation}\label{cca-ssg}\hspace*{-5pt}
\resizebox{.91\linewidth}{!}{$\displaystyle
\mathcal{L} = \Vert \mathbf{Z}^\text{\romannumeral1}-\mathbf{Z}^\text{\romannumeral2}\Vert^2_F + \lambda\underbrace{\left(\Vert {\mathbf{Z}^\text{\romannumeral1}}^\top\mathbf{Z}^\text{\romannumeral1} - \mathbf{I}_d \Vert^2_F + \Vert {\mathbf{Z}^\text{\romannumeral2}}^\top\mathbf{Z}^\text{\romannumeral2} - \mathbf{I}_d \Vert^2_F \right)}_{\text{dimension discrimination}}
$}
\end{equation}
Future improvements mainly focus on the effectiveness of the regularization term, such as how to avoid the global or local representation collapse problem~\cite{VICReg,LogDet}. 
\\[5pt]
{\bf Discussion.}
For feature prediction, both traditional methods and the latest masked autoencoders tend to preserve shallow features rather than capture deeper semantic information.
Despite that node instance discrimination encourages the model to focus on deeper semantic information, 
it mainly focuses on node feature correlations while neglecting high-order structural knowledge, potentially leading to suboptimal performance in structural tasks like link prediction.
Compared to discrimination between instances, dimension discrimination enables augmentation-invariant feature learning by clarifying node dimensions. However, as noted in~\cite{CCA-SSG}, its benefits may diminish with smaller representation dimensions. 

\conditionalblue{
{\it Efficiency analysis.} 
The computational efficiency of node feature-based tasks primarily depends on the size of the feature matrix \smash{$\mathbf{X} \in \mathbb{R}^{n \times d}$} and the model architecture. 
The time complexity for the most common InfoNCE-based node instance discrimination is $\mathcal{O}(n^-nd)$, where $n^-$ is the number of negative samples for each node. The huge demand for negative samples results in a prohibitive worst-case cost $\mathcal{O}(n^2d)$. 
Similarly, the time complexity for the dimension discrimination term in \eqref{cca-ssg} is $\mathcal{O}(nd^2)$.
In contrast, feature prediction and bootstrapping-based methods scale linearly with both $n$ and $d$ ($\mathcal{O}(nd)$), making them more advantageous in terms of scalability. The lightweight linear decoder adopted by new-generation masked autoencoders can further reduce the computational cost~\cite{MAE}.
} 

\subsection{Node Properties} \label{3.2}

While Section~\ref{3.1} discusses pretext tasks that focus on node semantic information, another class of tasks focuses on node properties, which are crucial for understanding their structural roles within the graph. 

The {\it centrality} is a set of node properties that quantify the relative importance between nodes based on local structure. 
For example, the degree of a node $\text{deg}(\cdot)$ is a common measure of local connectivity, defined as the number of edges incident to that node.
There are various kinds of centrality measures including degree, 
closeness, 
betweenness, 
eigenvector, 
and PageRank~\cite{PageRank}. 
Predicting centralities is one of the most direct methods for capturing the structural importance of nodes.
Autoencoding methods such as NWR-GAE~\cite{NWR-GAE} and MaskGAE~\cite{MaskGAE} employ an MSE loss to predict the node degree: $\mathcal{L} = \mathbb{E}_{i \in \mathcal{V}}[\Vert \text{deg}(i) - \hat{\text{deg}}(i)\Vert^2]$.
Besides degree prediction, CenPre~\cite{CenPre} leverages the left singular vector of $\mathbf{A}$ as structural representations, similar to the eigenvector centrality, and computes its similarity with the original features.
Hu {\it et al.}~\cite{UPGCN} propose {\it centrality ranking}, which predicts if a node has a higher or lower centrality score $s$ compared to another node.

There are also other node properties that capture structural features beyond relative importance. 
For instance, {\it (local) clustering coefficient}~\cite{clustercoef} measures the gathering tendency of node groups,
and predicting the clustering coefficient highlights the local relationship between nodes and guides the model to preserve them~\cite{SelfTask}.
{\it Node order} is also a special kind of property that holds the permutation-invariant information for more expressive GNNs.
PIGAE~\cite{PIGAE} aligns the order of output node representations with the input node order by adding a learnable permuter into a VGAE~\cite{GAE/VGAE}.
\\[5pt]
{\bf Discussion.} 
Node properties help models comprehend node information by encoding structural roles such as relative importance. 
However, node properties tend to be more task-specific~\cite{SelfTask}, meaning that they can be difficult to leverage as generalizable knowledge in certain scenarios. 
Additionally, node properties may not always provide sufficient discriminative power. Graphs with different topologies can have the same degree distribution, making it difficult to distinguish between them solely by the degree. 

\conditionalblue{
{\it Efficiency analysis.} 
Calculating different node properties necessitates varying degrees of additional computational costs. 
For example, the complexity for calculating the degree of all nodes is $\mathcal{O}(|\mathcal{E}|)$, while the complexity for calculating the eigenvector centrality (via eigenvalue decomposition) is $\mathcal{O}(n^3)$. 
Considering that most of these properties are scalars, the memory required to store them is usually $\mathcal{O}(n)$.
} 

\subsection{Links} \label{3.3}

From chemical bonds in molecular graphs and social connections in social networks to semantic relationships in knowledge graphs,
links play a fundamental role in graphs as they represent basic relationships between nodes.

{\it Link prediction} is a fundamental task in graph-based SSL aiming to predict the existence or probability of a link between two nodes. 
Structure-based autoencoders such as GAE~\cite{GAE/VGAE} feed the learned node representations into a dot-product decoder \smash{$\hat{\mathbf{A}}=\sigma(\mathbf{Z}\mathbf{Z}^\top)$} to predict the existence probability between a pair of nodes:
\begin{equation}\label{gae}
\mathcal{L} = -\mathbb{E}_{(i,j) \in \mathcal{E}}[\log \hat{A}_{i,j}] + \mathbb{E}_{(i,j) \in \mathcal{E}^-}[\log (1-\hat{A}_{i,j})]
\end{equation}
To capture more complicated latent spaces, 
variational autoencoders such as VGAE~\cite{GAE/VGAE} and \cite{ARGA/ARVGA,SIG-VAE,D-VGAE} learn a Gaussian model for latent embeddings $q(\mathbf{Z}|\mu, \sigma) = \mathcal{N}(\mu, \sigma^2)$ 
to approximate the real posterior $p(\mathbf{Z \vert \mathbf{X},\mathbf{A}})$.
Representation vectors are then sampled from these distributions to maximize the expected log-likelihood $\log p(\mathbf{A})$ bounded by the evidence lower bound (ELBO):
\begin{equation}\label{vgae}\displaystyle
\mathcal{J} = \mathbb{E}_{q(\mathbf{Z}|\mu, \sigma)}[\log p(\mathbf{A}\vert\mathbf{Z})] - D_{\text{KL}}[q(\mathbf{Z}|\mu, \sigma) \Vert p(\mathbf{Z})]
\end{equation}
where $p(\mathbf{Z}) =  \mathcal{N}(0, \mathbf{I})$ is the preset Gaussian prior.
Link prediction usually serves as an auxiliary task for training semi-supervised GNNs~\cite{SuperGAT}, discimination models~\cite{CG3,MVMI-FT,AEGCL}, 
and feature-based autoencoders~\cite{DLR-GAE,ASD-VAE}. 
Instead of a learnable decoder, discrimination losses such as InfoNCE are also applicable to link prediction~\cite{GraphFormers,Patton,HGPROMPT}. 

Another kind of prediction task, {\it link denoising}, involves adding random noises to the adjacency matrix.
For example, Bandana~\cite{Bandana} samples continuous edge noises and predicts the noise values. 
A more widely adopted approach for link denoising is 
{\it masked link prediction}, where a portion $p$ of edges is randomly masked using binary noise $M_{i,j} \sim Bernoulli(1-p)$. 
The objective is similar to that of binary link prediction, 
with the key difference being that only the masked edges are treated as positive samples during training. 
EdgeMask~\cite{SelfTask} and S2GAE~\cite{S2GAE} learn a decoder \smash{$\hat{\mathbf{A}}=\sigma(g(\mathbf{Z}\mathbf{Z}^\top;\Psi))$} to recover the masked edges. 
MaskGAE~\cite{MaskGAE} captures long-range relationships by randomly masking out paths obtained from random walks. 

Edge features in some attributed graphs also provide rich semantics complementing the graph structure, such as the number of co-authored papers or research topics in a co-authorship graph. 
Methods for node feature learning, such as auto-encoding in PIGAE~\cite{PIGAE} and ASD-VAE~\cite{ASD-VAE}, as well as masked feature prediction in AttrMask~\cite{GNNpretrain}, can be effortlessly applied to {\it edge feature prediction}.
\\[5pt]
{\bf Discussion.}
Link prediction has brought significant benefits to structure-based downstream tasks by capturing the structural information of graphs. 
It explicitly models the relationships between nodes that are not considered in node feature-based methods. 
Despite its widespread use, link prediction has been criticized for overemphasizing local structure~\cite{DGI,GraphMAE}. This highlights the need for exploring more compatible link-based pre-training strategies with feature semantics and higher-order structures. 

\conditionalblue{
{\it Efficiency analysis.} 
The memory required for dense and sparse adjacency matrices is $\mathcal{O}(n^2)$ and $\mathcal{O}(\vert\mathcal{E}\vert)$, respectively. 
Link prediction involves calculating the node similarity, such as the dot product and cosine similarity, both of which have a time complexity of $\mathcal{O}(d\vert\mathcal{E}\vert)$. 
As the efficiency of link prediction models is primarily bottlenecked by the edge size, pruning methods and masking link prediction can significantly reduce the memory cost for encoding.
} 

\section{Mesoscopic Pre-training Tasks} \label{4}

In contrast to microscopic tasks that focus on individual nodes and links, mesoscopic pre-training tasks aim to capture properties within a local range. These pretexts learn representations that encode higher-order information and long-range dependencies. 

\subsection{Context} \label{4.1}

Graph context refers to the neighborhood or a broader subgraph surrounding a node. 
Most context-based methods leverage the {\it homophily assumption}~\cite{homophily}, where adjacent nodes tend to share similar attributes. 

{\it Context discrimination} is a pretext that can be traced back to network embedding algorithms, 
e.g., DeepWalk~\cite{DeepWalk}. They sample node sequences from the graph using random walks and then iteratively update their embeddings using text embedding methods. 
GraphSAGE~\cite{GraphSAGE} redefines ``context'' from random walk sequences to the neighborhood subgraphs induced from the graph. 
It optimizes a negative sampling-based JS estimator loss \eqref{js} between the central node $i$ and its contextual nodes $j\in\mathcal{N}_i$:
\begin{equation}\label{graphsage}\hspace*{-5pt}
\resizebox{0.91\linewidth}{!}{$\displaystyle
\mathcal{L} = -\mathbb{E}_{\substack{i\in\mathcal{V}\\ j\in\mathcal{N}_i}}\big[\!\log \sigma(\boldsymbol{Z}_i^\top\boldsymbol{Z}_j) + \lambda\sum_{k\in \mathcal{V}^-}{\!\log \sigma (-\boldsymbol{Z}_i^\top\boldsymbol{Z}_k)}\big]
$}
\end{equation}
Later efforts~\cite{EGI,DGSI} 
improve GraphSAGE with a discriminator network to determine whether one node is the neighbor of another node.
COLES~\cite{COLES} captures neighborhood similarity by equipping Laplacian Eigenmaps~\cite{LE} with negative sampling, 
which is further generalized by GLEN~\cite{GLEN} as a rank optimization problem of representation scatter matrices. 

For other MI estimators, 
Graph-MLP~\cite{Graph-MLP} and 
N2N~\cite{N2N} define their positive sample pairs as every node and its $k$-hop neighborhood, and employ an InfoNCE loss.
Subg-Con~\cite{Subg-Con} selects $k$-nearest neighbors by personalized PageRank scores as positive samples of a triplet loss.
AFGRL~\cite{AFGRL} selects $k$-nearest neighbors in the context of both structure and feature as positive samples of a bootstrapping loss. 
HGRL~\cite{HGRL} further leverages the homophily assumption by selecting homophilic neighbors as precise positive samples. 
BSG~\cite{BSG} uses mean pooling to obtain neighborhood embeddings $\mathbf{Z}_\mathcal{N}$ and maximizes the mutual information $I(\mathbf{Z},\mathbf{Z}_\mathcal{N})$ and the conditional entropy $H(\mathbf{Z}\vert\mathbf{Z}_\mathcal{N})$ through MSE and hinge loss, respectively.

\begin{figure}[]
  \centering
\includegraphics[scale=0.85]{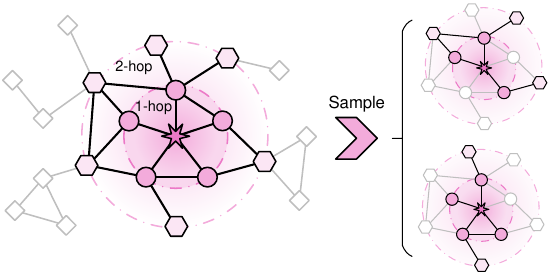}
\caption{ 
An illustration of the contextual knowledge. For the central node~\smash{\raisebox{-2pt}{\includegraphics[scale=0.6]{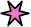}}}~of a 2-hop subgraph (Left), 
context discrimination often takes its neighboring nodes~\smash{\raisebox{-1pt}{\includegraphics[scale=0.7]{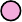}
}}~or~\smash{\raisebox{-1pt}{\includegraphics[scale=0.7]{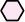}}}~as positive samples and other nodes~\smash{\raisebox{-1pt}{\includegraphics[scale=0.7]{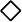}
}}~as negative samples. Contextual subgraph discrimination samples multiple contextual subgraphs (Right) as positive pairs, while negative ones are sampled from other subgraphs.
}
\label{context}
\end{figure}

Another task, {\it contextual subgraph discrimination}, measures the similarity between two different sampled subgraphs, as shown in Fig.~\ref{context}. 
ContextPred~\cite{GNNpretrain} samples a ``context graph'' from the periphery of the $k$-hop subgraph and matches them as a positive pair.
Instead of sampling an additional context graph,
GCC~\cite{GCC} directly induces two different subgraphs from the $k$-hop neighborhood of each node as a positive pair. 
S\textsuperscript{3}-CL~\cite{S3-CL} contrasts between intermediate message-passing layers to aggregate neighborhoods of varying scales. 
\\[5pt]
{\bf Discussion.} 
Compared to individual links, treating node context as instances facilitates a more complete understanding of the local graph structure. 
Nonetheless, it is empirically verified that some context learning methods have limited contributions to the performance of message-passing GNNs~\cite{SelfTask}, owing to the inherent capability of message-passing to extract local structural information. 
Context learning has the potential to benefit models that put more emphasis on global interactions, e.g., graph Transformers.

\conditionalblue{
{\it Efficiency analysis.} 
Using graph traversal algorithms, one can obtain the $k$-hop subgraph of any node with a time complexity of $O(n + |\mathcal{E}|)$. A more common approach is to aggregate the embeddings of neighboring nodes by left-multiplying $\mathbf{A}$, which has a time complexity of $O(k|\mathcal{E}|)$. Since the complexity is independent of $n$, this approach is particularly suitable for handling sparse networks with a large number of nodes. Existing GFM researchers~\cite{GraphFormers,PRODIGY,AllinOne,OFA} often reformulate node classification on large networks as predicting subgraph labels around the target node, which is essentially a divide-and-conquer strategy.
} 

\subsection{Long-range Similarities} \label{4.2}

Long-range similarities capture relationships between non-neighboring nodes that share semantic relevance. These similarities reveal higher-order dependencies beyond local neighborhoods. 
For example, in social networks, the small-world property implies that any two individuals are likely connected through a short chain of acquaintances~\cite{clustercoef}. 
\\[5pt]
{\bf Similarity prediction.} \label{4.2.1}
Similarity prediction aims to capture long-range similarities between nodes by directly predicting the similarity matrix $\mathbf{S} \in \mathbb{R}^{n \times n}$. 
Depending on whether two nodes are connected by a path, long-range similarities can be divided into topologically accessible similarities and topologically inaccessible ones. 
For the former, the {\it shortest path distance} measures the minimal distance between two connected nodes, and the {\it Katz index}~\cite{Katz} measures the total number of paths of every length between two connected nodes. 
S\textsuperscript{2}GRL~\cite{S2GRL} and PairwiseDistance~\cite{SelfTask} train the graph model to predict these similarities between all pairs of nodes by a negative log-likelihood loss.
For the latter, the {\it PageRank similarity}~\cite{PageRank} quantifies the importance of node pairs in terms of graph structure, while the {\it Jaccard's coefficient}~\cite{Jaccard} measures the overlap between node neighborhoods. 
There are also feature-based measures that quantify the degree of similarity between two nodes' features, regardless of their connectivity, such as Euclidean distance and cosine similarity~\cite{AGE,SelfTask,ULTRA-DP}. 
While Graph-Bert~\cite{Graph-Bert} directly predicts them by a regressive loss, 
AGE~\cite{AGE} and PairwiseAttrSim~\cite{SelfTask} adopt {\it similarity-based discrimination} that selects a subset of node pairs with the highest (resp. lowest) similarity scores and uses them as positive (resp. negative) samples. 
These similarities can bridge the disconnected components in the graph data which message-passing cannot.
\\[5pt]
\conditionalblue{
{\bf Similarity graph alignment.} \label{4.2.2}
Similarity graphs, derived from the original graph based on the pairwise similarities between nodes, usually serve as an alternative structural view of the graph. 
For instance, the kNN graph reconstructs the edge set by connecting the $k$-nearest neighbors of each node through various feature-based measures. 
It shares both commonalities and differences with the original graph structure (and other similarity graphs); therefore, aligning their semantics becomes a principled approach to combine feature semantics and graph structure. 
AM-GCN~\cite{AM-GCN} and DLR-GAE~\cite{DLR-GAE} minimize the discrepancy between original and similarity graph representations by MSE and cross-entropy. 
Instance discrimination methods~\cite{ASP,MVMI-FT,AEGCL} treat the original and similarity graphs as two views to integrate complementary information from both views. 
} 
\\[5pt]
{\bf Discussion.} 
Long-range similarities play a crucial role in capturing the dependencies between nodes out of reach for local contexts. 
It also enables the model to handle sparse graphs or graphs with disconnected components.  
However, the discrepancy between feature similarity and structural similarity can lead to semantic conflicts. 
Nodes with similar features may not always have similar structural neighborhoods. 
Therefore, the choice of similarity measures should depend on the real-world requirements. 

\conditionalblue{
{\it Efficiency analysis.} 
Computing all-pairs shortest paths can be computationally expensive for large graphs. 
The running time of the classic Floyd-Warshall algorithm is $\mathcal{O}(n^3)$ and the memory required is $\mathcal{O}(n^2)$. Using Dijkstra's algorithm for all nodes results in a complexity of $\mathcal{O}(n|\mathcal{E}|\log n)$. 
By contrast, calculating feature-based similarities requires $\mathcal{O}(n^2d)$ time, which is more efficient for dense networks. 
} 

\subsection{Motifs} \label{4.3}

Motifs are small subgraphs that frequently appear and carry significant structural and functional information, such as functional groups in molecular graphs, coregulators in regulatory networks, and cliques of people in social networks. 

\begin{figure}[]
  \centering
\includegraphics[scale=1.05]{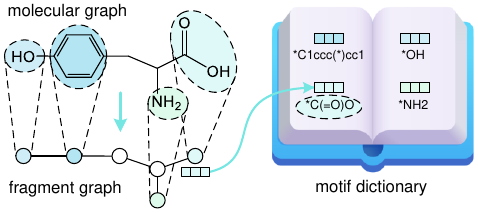}
\caption{
A molecular graph is converted into a fragment graph by aggregating functional groups into supernodes. For motif prediction, each supernode embedding is matched with a prototype in a motif dictionary. 
}
\label{fragment}
\end{figure}

{\it Motif prediction} tasks aim to learn motif-level representations by predicting the motif pseudo-labels of subgraphs. These pseudo-labels are given by unsupervised motif discovery algorithms, e.g., RDKit~\cite{RDKit}. 
GROVER~\cite{GROVER} assigns motif pseudo-labels to molecular graphs and trains a GNN for classification, and MoAMa~\cite{MoAMa} extends this idea by conducting motif-wise feature masking and prediction. 
DGPM~\cite{DGPM} performs a binary node-motif matching task to predict if a node belongs to a motif. 
Recent literature introduces the concept of ``fragment graphs'', 
whose nodes are aggregated from subgraphs containing specific motifs, shown in Fig.~\ref{fragment}.
The aggregated supernode representations are collected in a motif dictionary. In this way, motif prediction is transformed into a lookup task: the representation vector of each node is associated with an entry in the motif dictionary. 
MGSSL~\cite{MGSSL} proposed an autoregressive method to generate and classify the supernodes sequentially, while GraphFP~\cite{GraphFP} performs multi-label classification on the entire graph. 

Another line of work employs {\it motif-based discrimination} that creates contrastive sample pairs for motif-aware representations.
MotifRGC~\cite{MotifRGC} designs an adversarial motif generator to generate positive and negative views.
Fragment graphs can also serve as contrastive views: 
MICRO-Graph~\cite{MICRO-Graph} and GraphFP~\cite{GraphFP} treat the original graph and its corresponding fragment graph as a positive pair. 
\\[5pt]
{\bf Discussion.}
Most motif-based pretexts are designed specifically for molecular graphs, limiting their applicability to larger-scale networks. 
The only exception as far as we know is CTAug~\cite{CTAug}, a contrastive method aiming to preserve cohesive motifs (k-cliques, k-cores, etc.) in social networks. 
Future research should focus on developing more general motif learning methods 
to reduce the size of motif dictionaries while preserving essential structural and functional information. 

\conditionalblue{
{\it Efficiency analysis.} 
The efficiency of motif learning depends on the algorithm for obtaining motif pseudo-labels. For example, the IFG algorithm~\cite{IFG} called by RDKit has a time complexity $\mathcal{O}(n+\vert\mathcal{E}\vert)$ for every molecule.
Fragmentation methods traverse all nodes within a graph and map them to supernodes with extra $\mathcal{O}(n)$ time. However, it is worth noting that the diverse range of motifs can lead to large motif dictionaries, incurring extra memory overhead. 
}

\subsection{Clusters} \label{4.4}

\begin{figure}[]
  \centering
\includegraphics[scale=1.1]{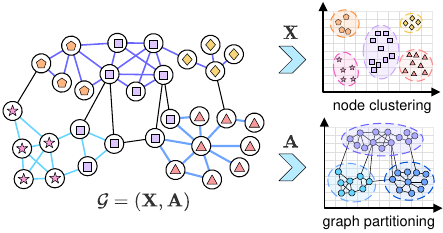}
\caption{ 
Node clustering and graph partitioning. The former clusters nodes mainly by feature similarity on $\mathbf{X}$, marked by node shapes and colors. The latter clusters nodes mainly by connection density on $\mathbf{A}$, marked by edge colors.
}
\label{cluster}
\end{figure}

Cluster-based tasks aim to learn representations that capture the inherent clustering structure of the graph, which can be defined based on either feature similarities or link connectivity. 
Clusters often possess a larger scale compared to motifs, providing a higher-level view of the graph structure. 
\\[5pt]
{\bf Node clustering.} \label{4.4.1}
Node clustering, a classic unsupervised learning task, is introduced as a pretext task by M3S~\cite{M3S} and NodeCluster~\cite{SS-GCN}. 
They leverage feature-based clustering algorithms (e.g., $K$-means~\cite{KmeansRL}, DeepCluster~\cite{DeepCluster}) to assign a cluster pseudo-label to every node. 
HomoGCL~\cite{HomoGCL} and MGSE~\cite{MGSE} first generate a prototype vector for every cluster by feature aggregation. 
Then, they optimize an MSE and a cross-entropy-based divergence loss of the cluster assignment probabilities, respectively. 
CARL-G~\cite{CARL-G} predicts ``cluster validation indices'', a set of measures indicating the compactness and separation of clusters. 
CommDGI~\cite{CommDGI}, S$^3$-CL~\cite{S3-CL} and more~\cite{HomoGCL,DCGL} 
focus on {\it cluster-based discrimination} where node embeddings are contrasted with learnable cluster prototypes. 
GraphLoG~\cite{GraphLoG} models the hierarchical nature of clustering by setting prototypes at different levels and organizing them in a tree structure.
\\[5pt]
{\bf Graph partitioning.} \label{4.4.2}
Graph partitioning is also known as ``non-overlapping community detection'' in the scenario of social network mining. Unlike node clustering, graph partitioning is based on structural community patterns and thus is available to unattributed graphs, illustrated in Fig.~\ref{cluster}.
Early works~\cite{UPGCN,SS-GCN} leverage unsupervised community detection methods, such as spectral clustering and Louvain~\cite{Louvain}, to generate partition pseudo-labels and learn a community indicator matrix. 
Distance2Clusters~\cite{SelfTask} performs a regression task between node representations and community prototypes. 
Several works have incorporated graph partitioning into more complex frameworks, e.g., link prediction VGAE~\cite{DGVAE} and masked autoencoders~\cite{Mask-GVAE}. 
%
{\it Partition-based discrimination} has also gained attention in recent works. 
gCooL~\cite{gCooL} enlarges the positive set by intra-community instances between two views, while CSGCL~\cite{CSGCL} uses the modularity-based community strength to weight node samples.  StructComp~\cite{StructComp} takes a different approach by compressing features of nodes in the same community and performs community-wise contrast with compressed features.
\\[5pt]
{\bf Discussion.}
Cluster-based pretexts provide graph models with a deeper understanding of the higher-level structural organization.
However, most cluster-based pretexts rely on non-overlapping algorithms, assuming that each node belongs to a single cluster. In real-world scenarios, nodes often belong to multiple overlapping communities, which remains a challenge for existing graph models.

\conditionalblue{
{\it Efficiency analysis.} 
The computational cost of clustering/partitioning algorithms can become prohibitive for large networks. 
Common implementations of $K$-means have the time complexity of $\mathcal{O}(Knd)$ for every iteration. 
For graph partitioning, vanilla spectral clustering reaches $\mathcal{O}(n^3)$ time complexity, while the modularity-based algorithms such as Louvain~\cite{Louvain} are believed to run in $\mathcal{O}(\vert\mathcal{E}\vert)$ time.
In practice, however, they still spend a lot of time processing large-scale networks, so clustering methods need to be considered carefully.
} 

\section{Macroscopic Pre-training Tasks} \label{5}

Unlike mesoscopic tasks that focus on the local graph structure, macroscopic pre-training tasks aim to capture global patterns and structures that span the entire graph. These pretexts are designed for a broader understanding of the overall organization and dynamics of the graph. 

\subsection{Global Structure} \label{5.1}

The goal of global structure-based tasks is to capture the overall topology and properties of a graph by learning from its global representations. 
\\[5pt]
{\bf Graph instance discrimination.} \label{5.1.1}
This task learns to distinguish between graph instances by focusing on graph-level representations. They are obtained by aggregating node embeddings by a simple readout function, such as mean pooling and summation.
GraphCL~\cite{GraphCL} matches positive and negative sample pairs from batches of small graphs with an InfoNCE estimator \eqref{infonce}, 
similar to node instance discrimination.
Other MI estimators are also suitable for graph instances, such as triplet loss~\cite{Mole-BERT} and bootstrapping loss~\cite{SimGRACE}. 
Subsequent works have explored various aspects of graph instance discrimination,
such as adaptive augmentations~\cite{JOAO,AD-GCL} 
and negative sample mining~\cite{CGC}. 

Graph representations can also be used to perform {\it node-graph discrimination}, also known as cross-scale contrast~\cite{TKDE22survey}. 
Here the global representation can form positive pairs with every node in the graph, and
negative pairs are generated by applying one-sided perturbations to either the node or the graph representation.
The JS estimator~\eqref{js} is in widespread use here, pioneered by DGI~\cite{DGI} and InfoGraph~\cite{InfoGraph}.
MVGRL~\cite{MVGRL} performs cross-view contrast by both node-level and graph-level perturbations. 
GGD~\cite{GGD} and D-SLA~\cite{D-SLA} propose {\it group discrimination}, a simplified binary classification approach predicting whether an instance belongs to the original or the perturbed view. 
This simplification greatly improves the efficiency, as calculating similarities between graph instances is no longer needed.
Node-graph discrimination captures the relationships between global and local representations of a graph, making it applicable to both small and large graphs~\cite{DGSI,SPAN}. 
\\[5pt]
{\bf Graph similarity prediction.} \label{5.1.2}
This pretext task leverages various kinds of graph-level similarity functions to learn graph-level representations, first envisioned by~\cite{GNNpretrain}. 
KernelPred~\cite{KernelPT} predicts various graph kernels, including the graphlet kernel, random walk kernel, WL subtree kernel, etc. These kernels capture different aspects of graph similarity, such as structural similarity, node proximity, and subgraph patterns.
D-SLA~\cite{D-SLA} generates a perturbed graph by adding and removing edges and predicts the graph edit distance kernel, 
the number of edge modifying steps between the original and perturbed graphs.
HTML~\cite{HTML} predicts the isomorphic similarity between graphs based on the Jaccard coefficient.
\\[5pt]
{\bf Discussion.}
Global structure-based tasks offer a holistic view of the entire graph, capturing its overall topology and properties. 
This is particularly advantageous when dealing with small graphs or scenarios focusing on global properties, including tasks such as graph classification and graph regression. 
However, the readout functions used to generate global representations can be coarse-grained, potentially losing important structural information. 
Moreover, graph perturbations can have a significant impact on the global semantics of small graphs. 

\conditionalblue{
{\it Efficiency analysis.} 
Although graph instance discrimination methods define readout functions on the entire graph, their simple formulations imply acceptable computational costs (usually less than $\mathcal{O}(n)$). 
However, calculating graph kernels on large networks is not an easy task, which makes prediction tasks more constrained by the data size.
} 

\subsection{Manifolds} \label{5.2}

Manifolds are underlying global topological patterns that Euclidean spaces struggle to represent. Recent work explores embedding graphs into non-Euclidean manifolds, such as hyperbolic or spherical spaces, to better model hierarchical and tree-like structures of graph data. 

{\it Cross-manifold discrimination} creates contrastive views in different manifolds, thereby capturing the unique properties of each manifold and their relationships.
HGCL~\cite{HGCL} uses a pair of hyperbolic GNNs to encode views of the graph, and DSGC~\cite{DSGC} uses both Euclidean and hyperbolic GNNs to obtain views in both spaces. 
RiemannGFM~\cite{RiemannGFM} contrasts between a hyperbolic and a spherical space. 
SelfMGNN~\cite{SelfMGNN} embeds graphs into a product space that combines Euclidean, hyperbolic, and spherical spaces. It enables adaptive learning of the most suitable manifold for each graph based on its structural properties.
\conditionalblue{
For prediction tasks, 
HDM-GAE~\cite{HDM-GAE} performs masked feature/link prediction in the hyperbolic space.
Graph-JEPA~\cite{Graph-JEPA} first expresses graph representations as angle vectors in a unit hyperbola and predicts them by a smooth-$\ell_1$ loss. 
} 
\\[5pt]
{\bf Discussion.}
Manifold-based tasks offer a promising new direction by capturing complex geometric structures and hierarchical relationships that are difficult to represent in Euclidean spaces. 
There is ample room for exploration, such as investigating more general and flexible approaches for graph manifold learning and exploring the integration of different manifolds other than a product space. 

\conditionalblue{
{\it Efficiency analysis.} 
The time overhead can arise from the switching between different topological spaces. 
For example, the exponential and logarithmic maps in hyperbolic spaces require multiple calculations of vector norms and inverse trigonometric functions. Although these operations do not alter the upper bound of the algorithm complexity, they introduce additional computations that increase the actual wall-clock time.
Furthermore, storing representations in different spaces incurs additional memory costs.
} 

\section{Downstream Tuning} \label{6}

\begin{figure*}[h!]
  \centering
  \includegraphics[scale=0.57]{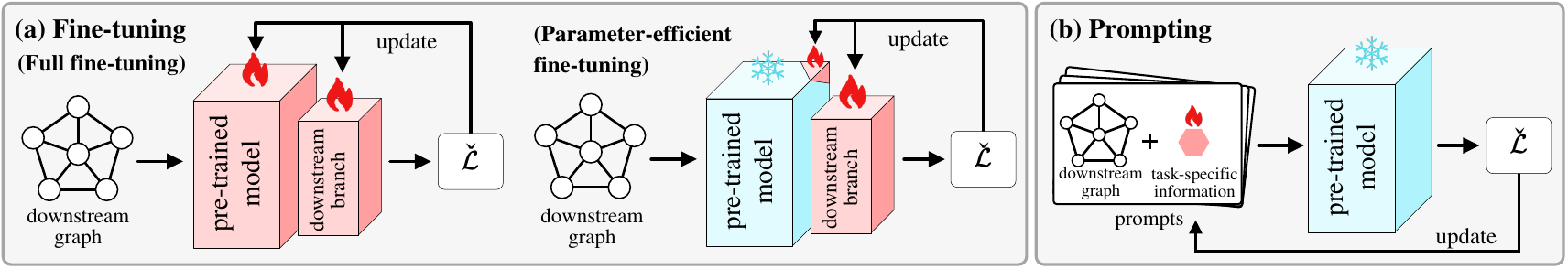}
  \caption{
  Different downstream tuning strategies. 
  \smash{\raisebox{-2pt}{\includegraphics[scale=0.35]{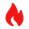}}}: tuned; \smash{\raisebox{-2pt}{\includegraphics[scale=0.35]{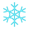}}}: frozen.
  \conditionalblue{
  (a) Graph fine-tuning, where the pre-trained parameters are updated along with downstream branches through the downstream task $\check{\mathcal{L}}$. 
  While full fine-tuning updates the entire set of pre-trained parameters, parameter-efficient fine-tuning only updates a small amount of it via specifically designed adapter modules.
  (b) Graph prompting, where the specifically designed graph prompts are updated, usually carrying downstream task-specific information. Tunable downstream branches are not necessary.
  } 
  }
  \label{downstream}
\end{figure*}

\tikzstyle{cyanleaf}=[draw=nodefeatureborderclr,
    rounded corners,minimum height=1em,
    fill=nodefeatureclr!40,text opacity=1, align=center,text=black,align=left,font=\scriptsize,
    inner xsep=3pt,
    inner ysep=1pt,
]
\tikzstyle{blueleaf}=[draw=nodepropertyborderclr,
    rounded corners,minimum height=1em,
    fill=nodepropertyclr!40,text opacity=1, align=center,text=black,align=left,font=\scriptsize,
    inner xsep=3pt,
    inner ysep=1pt,
]
\tikzstyle{purpleleaf}=[draw=linkborderclr,
    rounded corners,minimum height=1em,
    fill=linkclr!40,text opacity=1, align=center,text=black,align=left,font=\scriptsize,
    inner xsep=3pt,
    inner ysep=1pt,
]
\tikzstyle{pinkleaf}=[draw=contextborderclr,
    rounded corners,minimum height=1em,
    fill=contextclr!40,text opacity=1, align=center,text=black,align=left,font=\scriptsize,
    inner xsep=3pt,
    inner ysep=1pt,
]
\tikzstyle{redleaf}=[draw=longrangesimilarityborderclr,
    rounded corners,minimum height=1em,
    fill=longrangesimilarityclr!40,text opacity=1, align=center,text=black,align=left,font=\scriptsize,
    inner xsep=3pt,
    inner ysep=1pt,
]
\tikzstyle{orangeleaf}=[draw=motifborderclr,
    rounded corners,minimum height=1em,
    fill=motifclr!40,text opacity=1, align=center,text=black,align=left,font=\scriptsize,
    inner xsep=3pt,
    inner ysep=1pt,
]
\tikzstyle{yellowleaf}=[draw=clusterborderclr,
    rounded corners,minimum height=1em,
    fill=clusterclr!40,text opacity=1, align=center,text=black,align=left,font=\scriptsize,
    inner xsep=3pt,
    inner ysep=1pt,
]
\tikzstyle{limeleaf}=[draw=globalstructureborderclr,
    rounded corners,minimum height=1em,
    fill=globalstructureclr!40,text opacity=1, align=center,text=black,align=left,font=\scriptsize,
    inner xsep=3pt,
    inner ysep=1pt,
]
\tikzstyle{greenleaf}=[draw=manifoldborderclr,
    rounded corners,minimum height=1em,
    fill=manifoldclr!40,text opacity=1, align=center,text=black,align=left,font=\scriptsize,
    inner xsep=3pt,
    inner ysep=1pt,
]
\tikzstyle{promptmiddle}=[draw=microscopicclr,
    rounded corners,minimum height=1em,
    text opacity=1,align=center,text=black,align=left,font=\scriptsize,
    inner xsep=3pt,
    inner ysep=1pt,
]
\tikzstyle{finetunemiddle}=[draw=mesoscopicclr,
    rounded corners,minimum height=1em,
    text opacity=1,align=center,text=black,align=left,font=\scriptsize,
    inner xsep=3pt,
    inner ysep=1pt,
]

\tikzstyle{root}=[draw=black,
    rounded corners,minimum height=1em,
    fill=output-white!40,text opacity=1, align=center,
    fill opacity=.5,  text=black,align=left,font=\scriptsize,
    inner xsep=3pt,
    inner ysep=1pt,
]

\begin{figure*}[ht]
\centering
\begin{forest}
  for tree={
  forked edges,
  grow=east,
  reversed=true,
  anchor=base west,
  parent anchor=east,
  child anchor=west,
  base=middle,
  font=\scriptsize,
  rectangle,
  line width=0.7pt,
  draw=black,
  rounded corners,
  align=left,
  minimum width=2em,
  s sep=3pt,
  l sep=9pt,
  inner xsep=3pt,
  inner ysep=1pt,
  },
  where level=1{text width=4.5em}{},
  where level=2{text width=6em,font=\scriptsize}{},
  where level=3{font=\scriptsize}{},
  where level=4{font=\scriptsize}{},
  where level=5{font=\scriptsize}{},
  [Downstream tuning, root, rotate=90, anchor=north, edge=black
    [Graph fine-tuning\\(Section~\ref{6.1}), finetunemiddle, edge=longrangesimilarityborderclr,text width=6em
        [Full fine-tuning, redleaf, text width=7.3em, edge=longrangesimilarityborderclr
            [L2P-GNN~\cite{L2P-GNN}{,} S2PGNN~\cite{S2PGNN}{,} W2PGNN~\cite{W2PGNN}{,} G-Tuning~\cite{G-Tuning}{,} GraphControl~\cite{GraphControl}{,} GFT~\cite{GFT}, redleaf, text width=33.9em, edge=longrangesimilarityborderclr
            ]
            [Task similarity-based, redleaf, text width=7em, edge=longrangesimilarityborderclr
                [AUX-TS~\cite{AUX-TS}{,} GTOT-Tuning~\cite{GTOT}{,} Bridge-Tune~\cite{Bridge-Tune}, redleaf, text width=25.25em, edge=longrangesimilarityborderclr
                ]
            ]
        ]
        [Parameter-efficient\\fine-tuning, redleaf, text width=7.3em, edge=longrangesimilarityborderclr
            [AdapterGNN~\cite{AdapterGNN}{,} G-Adapter~\cite{G-Adapter}{,} GraphPAR~\cite{GraphPAR}{,} HG-Adapter~\cite{HG-Adapter}{,} DAGPrompT~\cite{DAGPrompT}, redleaf, text width=33.9em, edge=longrangesimilarityborderclr
            ]
        ]
    ]
    [Graph Prompting\\(Section~\ref{6.2}), promptmiddle, edge=nodefeatureborderclr, text width=6em
        [Node feature prompts, cyanleaf, text width=7.3em, edge=nodefeatureborderclr
            [GPF~\cite{GPF}{,} IA-GPL~\cite{IA-GPL}{,} DeepGPT~\cite{DeepGPT}, cyanleaf, text width=33.9em, edge=nodefeatureborderclr]
        ]
        [Class \& type prompts, cyanleaf, text width=7.3em, edge=nodefeatureborderclr
            [GPPT~\cite{GPPT}{,} PSP~\cite{PSP}{,} VNT~\cite{VNT}{,} SGL-PT~\cite{SGL-PT}{,} HGPROMPT~\cite{HGPROMPT}{,} HetGPT~\cite{HetGPT}{,} DAGPrompT~\cite{DAGPrompT}, cyanleaf, text width=33.9em, edge=nodefeatureborderclr]
        ]
        [Link prompts, cyanleaf, text width=7.3em, edge=nodefeatureborderclr
            [All in One~\cite{AllinOne}{,} CMRP~\cite{CMRP}{,} EdgePrompt~\cite{EdgePrompt}{,} FPrompt~\cite{FPrompt}, cyanleaf, text width=33.9em, edge=nodefeatureborderclr]
        ]
        [Contextual prompts, cyanleaf, text width=7.3em, edge=nodefeatureborderclr
            [GraphPrompt~\cite{GraphPrompt}{,} GraphPrompt+~\cite{GraphPrompt+}{,} Self-Pro~\cite{Self-Pro}{,} SUPT~\cite{SUPT}{,} ProNoG~\cite{ProNoG}{,}             GCoT~\cite{GCoT}, cyanleaf, text width=33.9em, edge=nodefeatureborderclr]
            [Prompt graph-based, cyanleaf, text width=6.7em, edge=nodefeatureborderclr
                [PRODIGY~\cite{PRODIGY}{,} OFA~\cite{OFA}, cyanleaf, text width=25.55em, edge=nodefeatureborderclr]
            ]
        ]
        [Multi-task prompts (Section \ref{8.1}), cyanleaf, text width=11em, edge=nodefeatureborderclr
            [ULTRA-DP~\cite{ULTRA-DP}{,} MultiGPrompt~\cite{MultiGPrompt}, cyanleaf, text width=30.2em, edge=nodefeatureborderclr]
        ]
        [Other prompts, cyanleaf, text width=7.3em, edge=nodefeatureborderclr
            [P2TAG~\cite{P2TAG}{,} GraphPro~\cite{GraphPro}{,} IGAP~\cite{IGAP}{,} TGPT~\cite{TGPT}, cyanleaf, text width=33.9em, edge=nodefeatureborderclr]
        ]
    ]
  ]
\end{forest}
\caption{
\conditionalblue{
Our taxonomy of downstream tuning strategies with representative literature.
} 
}
\label{dstaxonomy}
\end{figure*}

Downstream tuning in self-supervised graph models focuses on transferring the knowledge learned from self-supervised pretexts to downstream tasks, as formalized in~\eqref{gssl}. This section explores two main approaches: graph fine-tuning and graph prompting, illustrated in Fig.~\ref{downstream} and \ref{dstaxonomy}. These approaches offer different ways to leverage the pre-trained graph model for specific applications.

\subsection{Graph Fine-tuning} \label{6.1}

Fine-tuning adapts pre-trained models to downstream tasks by jointly training them with a generally simple task-specific branch. 
Traditional GNN fine-tuning methods, known as {\it full fine-tuning}, update all pre-trained parameters. 
To improve knowledge transfer, advanced techniques have emerged. For instance, L2P-GNN~\cite{L2P-GNN} uses a meta-learning framework that divides the pre-training data into support and query sets, simulating the adaptation process during pre-training. 
S2PGNN~\cite{S2PGNN} decomposes fine-tuning into multiple function modules and dynamically identifies the optimal modules for different downstream tasks. 
W2PGNN~\cite{W2PGNN} and G-Tuning~\cite{G-Tuning} focus on cross-domain transferability by representing fine-tuning as finding a combination of graphon bases. They serve as different dimensions of fundamental transferable patterns across data spaces. 
GraphControl~\cite{GraphControl} incorporates a conditional control module to utilize downstream task-specific features effectively.
GFT~\cite{GFT} rearranges the input graph data as trees with a virtual root node encoding task-specific information. In this way, downstream task-relevant nodes serve as children of the root node, and their information can be aggregated for various prediction tasks.

A specific line of work quantifies the generalization gap by the {\it task similarity} between pre-training $\mathcal{L}$ and downstream tasks $\check{\mathcal{L}}$.
GTOT-Tuning~\cite{GTOT} models graph fine-tuning as an optimal transport problem and minimizes the masked Wasserstein distance between tasks. 
AUX-TS~\cite{AUX-TS} introduces gradient similarity $sim(\mathcal{L}, \check{\mathcal{L}}) = \langle\nabla_{\Theta}\mathcal{L}, \nabla_{\Theta}\check{\mathcal{L}}\rangle$ which measures the similarity of loss surfaces between two tasks. 
If the similarity is positive, it indicates that the optimization directions during the gradient descent are non-conflicting, so two tasks are similar; and vice versa. 
Bridge-Tune~\cite{Bridge-Tune} defines representation consistency, the similarity between pairwise node label distributions. A binary label is assigned to each pair of nodes determined by whether their pretext pseudo-labels (or downstream labels) are the same.

Fine-tuning large-scale models can be computationally expensive, and biases from downstream tasks may compromise generalizability. To address these challenges, recent works adopt {\it parameter-efficient fine-tuning} (PEFT) modules such as adapters~\cite{Adapter,LoRA}, which enable models to update only a small subset of pre-trained parameters during fine-tuning. 
AdapterGNN~\cite{AdapterGNN} and G-Adapter~\cite{G-Adapter} introduce adapter modules tailored for GNNs and graph Transformers, respectively. 
DAGPrompT~\cite{DAGPrompT} proposes the Graph Low-Rank Adaptation (GLoRA) module. Different from LoRA~\cite{LoRA}, both the weights $\mathbf{W}$ and the adjacency matrix $\mathbf{A}$ in GLoRA are decomposed into two low-rank projection matrices $\mathbf{P},\mathbf{Q} \in \mathbb{R}^{d \times r}$ and $\mathbf{P}_{\mathbf{A}},\mathbf{Q}_{\mathbf{A}} \in \mathbb{R}^{n \times 1}$.
\\[5pt]
\conditionalblue{
{\bf Discussion.} Considering the gap between general graph knowledge and domain-specific downstream knowledge, pre-training and fine-tuning are currently indispensable for building a general graph model. 
However, graph fine-tuning methods often resort to specific designs in terms of tuning processes and model architectures, limiting their universality across different downstream scenarios. 
Moreover, fine-tuning the pre-trained parameters may harm the generalization and expressive power of the pre-trained model. 
Despite that PEFT methods enable precise fine-tuning with minimal resource requirements, they are less common in fine-tuning pure GNNs as they are relatively small in size. 
} 

\conditionalblue{
\subsection{Graph Prompting} \label{6.2}

Prompting is an emerging downstream tuning strategy that has gained popularity with the rise of LLMs. 
In the graph domain, prompting jointly encodes downstream graph data and corresponding task-specific information as additional learnable components called ``prompts''. During downstream training, only the learnable part of the prompts is updated, while the pre-trained model remains frozen. 
To this end, graph prompts should be first integrated into the downstream data before downstream training by various means (addition~\cite{GPF}, element-wise multiplication~\cite{GraphPrompt}, concatenation~\cite{VNT}, weighted aggregation~\cite{AllinOne}, linear transformation~\cite{GraphPrompt}, etc.), depending on the form of prompts and specific downstream requirements.
However, unlike prompts in natural language that follow a deterministic form, graph prompts can take various shapes, increasing the difficulty of prompt design. 
The following will discuss several graph prompt designs from a knowledge-based perspective. 
\\[5pt]
{\bf Node feature prompts}. Node feature prompts are learnable vectors $\boldsymbol{p} \in \mathbb{R}^d$ that share the same size with the node features. Node feature prompts are simple, effective, and generalizable, serving as an important cornerstone of graph prompting. The fundamentality of node features enables their generalization across different data domains and downstream tasks, such as node classification, link prediction, and graph classification. 
GPF~\cite{GPF} simply adds the node feature prompt $\boldsymbol{p} \in \mathbb{R}^d$ to every row of the downstream feature matrix, formally $\check{\mathbf{X}} \leftarrow [\check{\boldsymbol{X}}_i + \boldsymbol{p}]_{i\in\mathcal{V}}$ to perform supervised fine-tuning. 
GPF uses a universal feature prompt for every node, while its variant GPF-plus~\cite{GPF} assigns an independent prompt $\boldsymbol{p}_i$ to each node generated by a set of learnable prompt bases. 
IA-GPL~\cite{IA-GPL} generates a feature prompt for every input node from its representation through a vector quantization-based network. 
Aside from feature prompts, DeepGPT~\cite{DeepGPT} prepends prefix prompts to every embedding fed into a pre-trained graph Transformer. 
\\[5pt]
{\bf Class \& type prompts}. 
Class prompts are prototype vectors aggregated from pre-trained node representations that share the same class: $\boldsymbol{p}_c=agg(\boldsymbol{Z}_i|i\in\mathcal{V},y_i=c)$. They are associated with downstream node classes. 
Downstream tasks such as node classification can be achieved by matching node representations with these class prompts with cross-entropy~\cite{GPPT,VNT} or InfoNCE \eqref{infonce}~\cite{PSP,SGL-PT}. The selection of objectives often depends on the form of pretexts.

GPPT~\cite{GPPT} first divides the original graph into different clusters using node clustering algorithms (Section~\ref{4.4}), and then defines a set of independent class prompts within each cluster. 
Unlike GPPT which constructs independent node-class prompt pairs, PSP~\cite{PSP} connects class prototypes to the original graph as virtual class nodes and fine-tunes them by contrastive learning.
VNT~\cite{VNT} employs meta-learning on graph data with virtual class nodes, where the class prompts for each target task are aggregated from source tasks through an attention mechanism, aiming to bridge the gap between the source and target domains. 
SGL-PT~\cite{SGL-PT} leverages masked feature prediction to fine-tune class prompts. The graph classification problem is transformed into the reconstruction of a masked virtual supernode, which serves as the global representation.
DAGPrompT~\cite{DAGPrompT} concatenates the embeddings as well as class prompts from every GNN layer, and performs InfoNCE-based similarity learning.
Type prompting-based methods, such as HGPROMPT~\cite{HGPROMPT} and HetGPT~\cite{HetGPT}, assign a prompt to each node type, similar to class prototypes. Unlike class prompts, node types are considered node properties that are common in heterogeneous graphs and typically do not require manual labeling.
\\[5pt]
{\bf Link prompts}. 
Considering the structural knowledge carried by downstream graph data is largely overlooked by node feature-based prompts, prompts on adjacency matrices begin to thrive. 
EdgePrompt~\cite{EdgePrompt} proposes edge prompts that are learnable attributes on every edge and are integrated into node representations through message propagation. Similar to GPF, edge prompts can be either one shared vector or customized vectors generated by prompt bases. 
All in One~\cite{AllinOne} considers pairwise relationships between prompt tokens and constructs a graph prompt $\mathcal{G}_{\boldsymbol{p}}=(\mathbf{X}_{\boldsymbol{p}},\mathbf{A}_{\boldsymbol{p}})$, where $\mathbf{X}_{\boldsymbol{p}} = [\boldsymbol{p}_k]_{k}$ and $\mathbf{A}_{\boldsymbol{p}} = [\langle\boldsymbol{p}_k,\boldsymbol{p}_l\rangle]_{k,l}$. A meta-learning strategy is developed to adapt All in One to miscellaneous downstream scenarios. 
\\[5pt]
{\bf Contextual prompts}.  
Context information is often considered in prompting methods in the form of aggregated neighborhood features or embeddings. 
GPPT~\cite{GPPT} and GraphPrompt series~\cite{GraphPrompt,GraphPrompt+} design structural prompts that encode one-hop aggregated contextual information for downstream tuning.
Self-Pro~\cite{Self-Pro} constructs a 2-hop adjacency matrix $\mathbf{A}_2 = [A_{i,j}=1 \cap A_{j,k}=1]_{i,k}$ as the contextual prompt. 
SUPT~\cite{SUPT} builds upon the learnable prompt bases in GPF-plus~\cite{GPF}. However, it uses a message-passing approach to aggregate these bases, preserving the semantic similarity among neighboring nodes in the prompts. 
ProNoG~\cite{ProNoG} generates contextual prompts by employing a condition-net, where the input representations are aggregated from $k$-hop subgraphs. 
Building on this, GCoT~\cite{GCoT} iteratively aggregates representations from every hidden layer of a pre-trained GNN as ``thoughts'', simulating the Chain-of-Thought reasoning in large language models.

Some prompting methods move away from feature vectors and seek other effective forms.
For example, PRODIGY~\cite{PRODIGY} and OFA~\cite{OFA} construct a prompt graph\footnote{Unlike the existing survey~\cite{PromptSurvey} which categorizes both All in One~\cite{AllinOne} and PRODIGY~\cite{PRODIGY} as ``Prompt as Graphs'', we explicitly distinguish them by different notions: ``{\it graph prompt}'', a graph added on the downstream graph as an entire prompt; and ``{\it prompt graph}'', a new graph comprised of prompt nodes and class nodes.}, in which each prompt node (data node) represents a sampled $k$-hop subgraph and each class node represents a class to which the central node belongs. Links between prompt nodes and class nodes indicate the task-specific supervision signals. 
Learning to predict these links has been demonstrated to be a highly generalizable strategy, applicable both to pre-training and fine-tuning to facilitate both few-shot in-context learning~\cite{PRODIGY} and zero-shot learning~\cite{OFA}. 
Besides, PRODIGY updates the prompt graph with an auxiliary self-supervised context prediction objective: it predicts if one node belongs to the $k$-hop subgraph of another target node.
\\[5pt]
{\bf Other prompts}.  
Some cutting-edge prompting methods explore underlying topological properties of graph data.
IGAP~\cite{IGAP} designs a spectral prompt to transform the low-dimensional pre-training domain to the fine-tuning domain, as the low-frequency domain describes local smooth patterns of graph signals. 
TGPT~\cite{TGPT} captures graphlet information -- small motifs that describe local structure patterns of a node -- into its node-level and the graph-level prompts. Specifically, node-level topology-aware prompts are generated from a fast graphlet transform matrix, and graph-level ones are aggregated from them.
\\[5pt]
{\bf Discussion.}
Graph prompting is a novel approach to envisioning graph downstream tuning.
It can be viewed as a form of PEFT on graph data instead of model architecture,
where prompts are considered tunable adapters across the pre-training and downstream data spaces, achieving both effectiveness and efficiency.
Prompting not only bridges data domains but also provides unified fine-tuning templates for various downstream prediction tasks. 
Such templates include link prediction~\cite{Self-Pro,PRODIGY,OFA,DAGPrompT}, graph classification~\cite{AllinOne}, subgraph similarity prediction~\cite{GraphPrompt,GraphPrompt+}, and more.
Despite the promising advancements, graph prompting remains a developing area, offering opportunities for further research and improvement. 
For instance, many graph prompts remain challenging for humans to comprehend, posing challenges for the explainability of graph prompting. 
Additionally, most proposed ``unified prompt templates for downstream tasks'' primarily focus on discriminative tasks such as link prediction and graph classification, while neglecting generative and other open-ended tasks that also have widespread demand. 
} 

\section{Self-supervised Graph Language Models} \label{7}

Previous sections have explored how self-supervised GFMs learn different types of graph knowledge through pre-training and downstream tuning. The emergence of large language models (LLMs) has opened up new avenues for constructing graph language models (GLMs) that leverage knowledge patterns, architectures, and training strategies from the natural language domain to process graph data. While traditional GFMs excel at capturing structural patterns, GLMs aim to bridge the gap between graph topology and semantic understanding by combining the strengths of both GNNs and language models.

This section examines self-supervised GLMs from two perspectives shown in Fig.~\ref{glmtaxonomy}: 
(1) pre-training GLMs with self-supervision, which focuses on incorporating graph knowledge into language modeling pre-training and graph pre-training of GLMs; and 
(2) tuning GLMs, which focuses on adapting pre-trained language knowledge to graph-specific scenarios through techniques like prompting and fine-tuning.
In what follows, we delve into these two directions, particularly focusing on how GLMs integrate different forms of graph knowledge.

\tikzstyle{cyanleaf}=[draw=nodefeatureborderclr,
    rounded corners,minimum height=1em,
    fill=nodefeatureclr!40,text opacity=1, align=center,text=black,align=left,font=\scriptsize,
    inner xsep=3pt,
    inner ysep=1pt,
]
\tikzstyle{blueleaf}=[draw=nodepropertyborderclr,
    rounded corners,minimum height=1em,
    fill=nodepropertyclr!40,text opacity=1, align=center,text=black,align=left,font=\scriptsize,
    inner xsep=3pt,
    inner ysep=1pt,
]
\tikzstyle{purpleleaf}=[draw=linkborderclr,
    rounded corners,minimum height=1em,
    fill=linkclr!40,text opacity=1, align=center,text=black,align=left,font=\scriptsize,
    inner xsep=3pt,
    inner ysep=1pt,
]
\tikzstyle{pinkleaf}=[draw=contextborderclr,
    rounded corners,minimum height=1em,
    fill=contextclr!40,text opacity=1, align=center,text=black,align=left,font=\scriptsize,
    inner xsep=3pt,
    inner ysep=1pt,
]
\tikzstyle{redleaf}=[draw=longrangesimilarityborderclr,
    rounded corners,minimum height=1em,
    fill=longrangesimilarityclr!40,text opacity=1, align=center,text=black,align=left,font=\scriptsize,
    inner xsep=3pt,
    inner ysep=1pt,
]
\tikzstyle{orangeleaf}=[draw=motifborderclr,
    rounded corners,minimum height=1em,
    fill=motifclr!40,text opacity=1, align=center,text=black,align=left,font=\scriptsize,
    inner xsep=3pt,
    inner ysep=1pt,
]
\tikzstyle{yellowleaf}=[draw=clusterborderclr,
    rounded corners,minimum height=1em,
    fill=clusterclr!40,text opacity=1, align=center,text=black,align=left,font=\scriptsize,
    inner xsep=3pt,
    inner ysep=1pt,
]
\tikzstyle{limeleaf}=[draw=globalstructureborderclr,
    rounded corners,minimum height=1em,
    fill=globalstructureclr!40,text opacity=1, align=center,text=black,align=left,font=\scriptsize,
    inner xsep=3pt,
    inner ysep=1pt,
]
\tikzstyle{greenleaf}=[draw=manifoldborderclr,
    rounded corners,minimum height=1em,
    fill=manifoldclr!40,text opacity=1, align=center,text=black,align=left,font=\scriptsize,
    inner xsep=3pt,
    inner ysep=1pt,
]
\tikzstyle{tunemiddle}=[draw=microscopicclr,
    rounded corners,minimum height=1em,
    text opacity=1,fill=nodepropertyclr!20,align=center,text=black,align=left,font=\scriptsize,
    inner xsep=3pt,
    inner ysep=1pt,
]
\tikzstyle{pretrainmiddle}=[draw=mesoscopicclr,
    rounded corners,minimum height=1em,
    text opacity=1, fill=longrangesimilarityclr!20,align=center,text=black,align=left,font=\scriptsize,
    inner xsep=3pt,
    inner ysep=1pt,
]

\tikzstyle{root}=[draw=black,
    rounded corners,minimum height=1em,
    fill=output-white!40,text opacity=1, align=center,
    fill opacity=.5,  text=black,align=left,font=\scriptsize,
    inner xsep=3pt,
    inner ysep=1pt,
]

\begin{figure*}[ht]
\centering
\captionsetup{aboveskip=3pt}
\begin{tikzpicture}
\node[anchor=north west] (tree) {
\begin{forest}
  for tree={
  forked edges,
  grow=east,
  reversed=true,
  anchor=base west,
  parent anchor=east,
  child anchor=west,
  base=middle,
  font=\scriptsize,
  rectangle,
  line width=0.7pt,
  draw=black,
  rounded corners,
  align=left,
  minimum width=2em,
  s sep=3pt,
  l sep=9pt,
  inner xsep=3pt,
  inner ysep=1pt,
  },
  where level=1{text width=4.5em}{},
  where level=2{text width=6em,font=\scriptsize}{},
  where level=3{font=\scriptsize}{},
  where level=4{font=\scriptsize}{},
  where level=5{font=\scriptsize}{},
  [Self-supervised GLMs, root, rotate=90, anchor=north, edge=black
    [Pre-training GLMs \\with self-supervision\\(Section~\ref{7.1}), pretrainmiddle, edge=longrangesimilarityborderclr,text width=8em
        [Language modeling pre-training, redleaf, text width=11em, edge=longrangesimilarityborderclr
            [Patton~\cite{Patton}{,} UniGraph~\cite{UniGraph}{,} P2TAG~\cite{P2TAG}{,} THLM~\cite{THLM}{,} Path-LLM~\cite{Path-LLM}, redleaf, text width=28.2em, edge=longrangesimilarityborderclr
            ]
        ]
        [Graph pre-training, redleaf, text width=6.6em, edge=longrangesimilarityborderclr
            [GraphFormers~\cite{GraphFormers}{,} Patton~\cite{Patton}{,} GALM~\cite{GALM}{,} G2P2~\cite{G2P2}{,} GRENADE~\cite{GRENADE}{,} ConGraT~\cite{ConGraT}{,} \\THLM~\cite{THLM}{,} 
            GraphGPT~\cite{GraphGPT}{,} UniGLM~\cite{UniGLM}{,} GSPT~\cite{GSPT}{,} LLaSA~\cite{LLaSA}, redleaf, text width=32.6em, edge=longrangesimilarityborderclr
            ]
        ]
    ]
    [Tuning GLMs\\(Section~\ref{7.2}), tunemiddle, edge=nodefeatureborderclr, text width=5em
        [Prompting for GLMs, cyanleaf, text width=7.3em, edge=nodefeatureborderclr
            [NLGraph~\cite{NLGraph}{,} Beyond Text~\cite{BeyondText}{,} GraphQA~\cite{TalkLikeAGraph}{,} Huang {\it et al.}~\cite{LLM-Structured-Data}{,} ProGraph~\cite{ProGraph}{,} SNS~\cite{SNS}{,} 
            \\Skianis {\it et al.}~\cite{Pseudo-Code}{,} AskGNN~\cite{AskGNN}{,} LLM4DyG~\cite{LLM4DyG}, cyanleaf, text width=34.9em, edge=nodefeatureborderclr]
        ]
        [Fine-tuning for GLMs, purpleleaf, text width=7.3em, edge=linkborderclr
            [Fine-tuning\\modules, purpleleaf, text width=4em, edge=linkborderclr
                [LLM-empowered\\representations, purpleleaf, text width=5.7em, edge=linkborderclr
                    [GraphToken~\cite{GraphToken}{,} GraphPrompter~\cite{GraphPrompter}{,} G-Prompt~\cite{G-Prompt}{,} \\ GraphAdapter~\cite{GraphAdapter}{,} DGTL~\cite{DGTL}{,} GOFA~\cite{GOFA}{,}  TAGA~\cite{TAGA}{,} \\GraphCLIP~\cite{GraphCLIP}{,} Pan {\it et al.}~\cite{DistillLLM}{,} ENGINE~\cite{ENGINE}{,} TEA-GLM~\cite{TEA-GLM}{,} \\GraphGPT~\cite{GraphGPT}{,} HiGPT~\cite{HiGPT}{,} LLaGA~\cite{LLaGA}{,} HIGHT~\cite{HIGHT}{,} \\GraphTranslator~\cite{GraphTranslator}, purpleleaf, text width=21.9em, edge=linkborderclr]
                ]
                [LLM-empowered\\augmentations, purpleleaf, text width=5.7em, edge=linkborderclr
                    [Text{:} TAPE~\cite{TAPE}{,} KEA~\cite{KEA}{,} LLM4Mol~\cite{LLM4Mol}{,} TANS~\cite{TANS}{,} \\\quad GAugLLM~\cite{GAugLLM}{,} SFGL~\cite{SFGL}, purpleleaf, text width=21.9em, edge=linkborderclr]
                    [Pseudo-labels{:} LLM-GNN~\cite{LLM-GNN}{,} 
                    LLM4RGNN~\cite{LLM4RGNN}{,} Locle~\cite{Locle}, purpleleaf, text width=21.9em, edge=linkborderclr]
                    [Graph structure{:} LLM4NG~\cite{LLM4NG}{,} OpenGraph~\cite{OpenGraph}{,} LOGIN~\cite{LOGIN}{,} \\\quad Sun {\it et al.}~\cite{A-D+LPA}
                    , purpleleaf, text width=21.9em, edge=linkborderclr]
                ]
            ]
            [Tuning LLM\\backbones, purpleleaf, text width=4.3em, edge=linkborderclr
                [GIANT~\cite{GIANT}{,} GLEM~\cite{GLEM}{,} GraphLLM~\cite{GraphLLM}{,} LEADING~\cite{LEADING}{,} InstructGLM~\cite{InstructGLM}{,} \\SimTeG~\cite{SimTeG}{,} WalkLM~\cite{WalkLM}{,} GUNDAM~\cite{GUNDAM}{,} LinguGKD~\cite{LinguGKD}{,} InstructGraph~\cite{InstructGraph}{,} \\GraphWiz~\cite{GraphWiz}{,}  GraphInstruct~\cite{GraphInstruct}{,} CMRP~\cite{CMRP}{,} AuGLM~\cite{AuGLM}{,} HierPromptLM~\cite{HierPromptLM}, purpleleaf, text width=28.95em, edge=linkborderclr]
            ]
        ]
    ]
  ]
\end{forest}
};
\node[anchor=north west, xshift=3.6cm, yshift=-18pt, opacity=1] at (tree.north west) {
\includegraphics[scale=0.5]{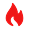}};
\node[anchor=north west, xshift=5.6cm, yshift=-1.5cm, opacity=1] at (tree.north west) {
\includegraphics[scale=0.5]{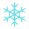}};
\node[anchor=north west, xshift=5.6cm, yshift=-4.8cm, opacity=1] at (tree.north west) {
\includegraphics[scale=0.5]{fig/icon/tunedtransparent.png}};
\end{tikzpicture}
\caption{
\conditionalblue{
Our taxonomy of self-supervised GLMs with representative literature. 
\smash{\raisebox{-2pt}{\includegraphics[scale=0.35]{fig/icon/tuned.png}}}: network parameter is updated; \smash{\raisebox{-2pt}{\includegraphics[scale=0.35]{fig/icon/frozen.png}}}: parameter frozen.
} 
}
\label{glmtaxonomy}
\end{figure*}

\conditionalblue{
\subsection{Pre-training GLMs with Self-supervision} \label{7.1}

GLM pre-training includes two distinct categories: {\it language modeling pre-training}, and {\it graph pre-training}. 
Note that both pre-training schemes can be applied to graph models such as GNNs and graph transformers, as well as to open-source language models (LMs) like T5~\cite{T5} and BERT~\cite{BERT}.

\subsubsection{Language Modeling Pre-training} \label{7.1.1}
Language modeling pre-training refers to a set of self-supervised pretext tasks in the language domain, known as ``language modeling''.
They mainly include {\it \textbf{a}uto\textbf{r}egressive language modeling} (AR) and {\it \textbf{m}asked \textbf{l}anguage \textbf{m}odeling} (MLM).
AR, also known as causal language modeling and next-token prediction, predicts the next token in a sequence usually by a maximum likelihood estimation loss. 
The well-known AR models include GPT series~\cite{GPT}, LLaMA~\cite{LLaMA}, and DeepSeek~\cite{DeepSeek-R1}. 
MLM, exemplified by BERT~\cite{BERT} and RoBERTa~\cite{RoBERTa}, predicts masked tokens using bidirectional context, making it ideal for understanding tasks but less suited for generation. 
Additionally, hybrid methods such as  T5~\cite{T5} and BART~\cite{BART} combine AR and MLM to handle both understanding and generation, though at the cost of higher complexity. 

Many early GLMs directly utilize open-source or closed-source LLMs pre-trained by the aforementioned tasks. 
Although pure LLMs have demonstrated preliminary abilities in handling and reasoning on graphs~\cite{NLGraph}, the modality gap between text and graphs makes graph tasks challenging for LLMs without additional support~\cite{BeyondText,TalkLikeAGraph}. 
Therefore, several attempts integrate GNNs as components of LLMs and jointly pre-train them by AR and MLM. 
UniGraph~\cite{UniGraph} and P2TAG~\cite{P2TAG} concatenate an LM and a GNN together and perform MLM on textual node attributes. THLM~\cite{THLM} jointly pre-trains a BERT and a heterogeneous GNN by MLM. 
On top of this, graph-oriented language pre-training methods are developed. 
Patton~\cite{Patton} improves traditional MLM to contextualized MLM: it utilizes both representations of the current node and its neighboring nodes to predict the missing tokens of the current node.
Path-LLM~\cite{Path-LLM} first generates textual sequences by interconnecting node text in the same order as the shortest path, where the earlier node text becomes the prefix tokens for AR pre-training. 

\subsubsection{Graph Pre-training} \label{7.1.2}

Grap pre-training features self-supervised pretext tasks in the graph domain, as introduced in Section~\ref{3}--\ref{5}. They are adapted to pre-train or assist in pre-training language model architectures. 
Cross-entropy-based and InfoNCE-based link prediction (Section~\ref{3.3}) become the first choice:
GALM~\cite{GALM} concatenates an LM and a GNN for joint pre-training, while GraphFormers~\cite{GraphFormers} and Patton~\cite{Patton} incorporate GNN and Transformer layers into a hybrid framework. 
UniGLM~\cite{UniGLM} employs node instance discrimination (Section~\ref{3.1}) to pre-train a language model, where positive node samples are selected from multiple hops of the central node by Personalized PageRank.

In contrast to single-modal discrimination, {\it node-text discrimination} is a more common approach inspired by existing multimodal pre-training models like CLIP~\cite{CLIP}. This approach discriminates between representations derived from a GNN and an LM/LLM to align the two modalities.
GraphGPT~\cite{GraphGPT} trains a textual and a graph Transformer by minimizing a cross-entropy-based discrimination loss to align their representation spaces for future instruction tuning. 
LLaSA~\cite{LLaSA} unifies different types of structured data into hypergraphs and performs contrastive learning between a hypergraph GNN and a hybrid Transformer.

To achieve more effective modality alignment, researchers agree on the necessity of incorporating higher-order graph-specific knowledge during the self-supervised pre-training. 
Graph context (Section~\ref{4.1}) becomes especially crucial for GLMs, 
as it is easy for LMs to understand compared to other structural knowledge, and provides a clearer perspective of the textual semantics of node entities and their adjacency relationships. 
G2P2~\cite{G2P2} jointly pre-trains a Transformer and a GCN by context-aware node-text discrimination: it generates a summary embedding by averaging neighborhood text embeddings and performs discrimination among node, text, and summary. 
GRENADE~\cite{GRENADE} employs both node-level and neighborhood-level discrimination within and between GNN and BERT. It minimizes the KL divergence of neighbor similarity distributions. 
THLM~\cite{THLM} introduces an auxiliary objective (along with MLM) to differentiate contextual and distant nodes.
Another common graph knowledge is long-range similarities (Section~\ref{4.2}).
ConGraT~\cite{ConGraT} jointly pre-trains a sentence Transformer and a GAT by node-text discrimination with a long-range similarity function, namely the number of common neighbors and SimRank. 
GSPT~\cite{GSPT} pre-trains a Transformer by masked feature prediction (Section~\ref{3.1}) on generated random walk sequences. 
They employ a node feature reconstruction loss rather than predicting the masked token IDs as in MLM. 
} 

\conditionalblue{
\subsection{Tuning GLMs} \label{7.2}

Downstream tuning is the dominant approach for deploying GLMs, given the high computational demands of LLM pre-training. 
LLMs to be tuned are usually pre-trained by large-scale language modeling pretexts, endowing them with a certain level of world knowledge. 
They can be utilized for either adapting to graph-specific downstream scenarios or guiding the training of other downstream branches. 
Some literature categorizes the roles of LLMs as enhancers, predictors, encoders, aligners, etc.~\cite{KEA,GLM-TKDE24,LLMsurveyIJCAI}, viewing pre-trained LLMs as auxiliary modules. 
Instead, we consider pre-trained LLMs as the core model and treat all subsequent training processes as downstream tuning. 
This perspective allows us to more clearly demonstrate how graph-specific knowledge assists LLMs in graph downstream tuning.

Note that GLMs discussed in this section do not necessarily employ self-supervised pre-training, and they may be designed for only one specific downstream task. 
However, here we focus solely on the tuning strategies employed, 
which we believe have the potential to be applied to self-supervised GLMs or to provide valuable insights.
} 

\subsubsection{Prompting for GLMs} \label{7.2.1}

With the advent of LLMs, the earliest GLMs attempt to directly utilize closed-source LLMs such as GPT-3.5 to construct graph learners. 
Specifically, by designing particular graph-specific prompts, LLMs can get a grasp of key graph knowledge for open-ended tasks such as question answering and reasoning. 
Some methods also rely on in-context learning~\cite{GraphText,AskGNN}, i.e., to provide a few task-specific examples within the prompts to guide the LLM output. 
Many early attempts, including empirical studies on using LLMs for graph tasks~\cite{NLGraph,BeyondText,TalkLikeAGraph,LLM-Structured-Data,ProGraph}, resort to graph-specific prompts rather than fine-tuning strategies. 

To achieve effective prompting, the first step is to convert graphs into text using natural language or structured language~\cite{GraphText,InstructGLM}. 
However, merely inputting textualized graphs is insufficient for LLMs to fully comprehend graph structural knowledge. Various methods have been proposed to further embed graph knowledge into prompts. 
For instance, SNS~\cite{SNS} ranks the textual similarity between nodes and selects the top-2 similar neighbors as additional instructions. 
Skianis {\it et al.}~\cite{Pseudo-Code} textualize the reasoning process of various graph question-answering tasks as pseudo-code functions.
AskGNN~\cite{AskGNN} employs a GNN to select an optimal set of examples for in-context learning. 

\subsubsection{Fine-tuning for GLMs} \label{7.2.2}

The rise of open-source LLMs, coupled with the limitations of prompting models in terms of cross-modal and zero-shot generalization capabilities, has facilitated the research of graph-specific fine-tuning for GLMs. 
{\it Instruction tuning} is the primary fine-tuning approach for LLMs, wherein appropriate queries for downstream tasks are constructed to elicit the desired LLM predictions, and model parameters are updated by minimizing an error function. 
However, LLMs typically possess a larger parameter scale than GNNs or GTs, making it challenging for GLMs to perform full fine-tuning. 
Consequently, parameter-efficient fine-tuning (PEFT) methods have become the predominant approach, where the LLM is frozen and a special fine-tuning module is tuned instead.
Apart from intrinsic adapters such as LoRA~\cite{LoRA}, GLMs often incorporate tailored fine-tuning modules to (1) maximize the retention of pre-trained LLMs' powerful semantic encoding ability, and (2) align the semantic spaces of text and graphs.
Types, interfaces, and tuning strategies of fine-tuning modules vary. 
\\[5pt]
{\bf Module types: GNNs vs. Non-GNNs.} The most commonly used fine-tuning module is GNN, which can capture structural knowledge of graphs, thereby helping to bridge the modality gap. 
The training of these modules can rely on both downstream task-relevant information and self-supervision signals.
GraphToken~\cite{GraphToken} and GraphPrompter~\cite{GraphPrompter} utilize a GNN to generate graph tokens as input for LLMs, and tune\footnote{By ``tuning'' we mean that the GNN serves as an auxiliary module of a pre-trained LLM. 
From the perspective of the GNN, it may be trained from scratch and regarded as pre-training in the original paper. This does not conflict with our statements in this section. }
the GNN by answering graph-related questions. 
Conversely, G-Prompt~\cite{G-Prompt} uses MLM to tune a GNN following an LM to obtain graph representations through textual prompts.
GraphAdapter~\cite{GraphAdapter} tunes a GNN by AR as a graph learning branch of a pre-trained LM. Then, it merges the representations with an MLP-based fusion block before passing them to a downstream head.
GOFA~\cite{GOFA} interleaves GNN layers into a pre-trained LLM encoder as adapters and employs various self-supervised fine-tuning methods, including AR, shortest path distance prediction, and common neighbor prediction. 
DGTL~\cite{DGTL} first generates disentangled graph embeddings by assigning diverse edge weights to each GNN layer. These embeddings are then incorporated into the input tokens of each LLM layer to perform context-aware instruction tuning. 
TAGA~\cite{TAGA} and GraphCLIP~\cite{GraphCLIP} employ node-text discrimination fine-tuning to align graph and text. 
TEA-GLM~\cite{TEA-GLM} tunes a GNN using both node instance and dimension discrimination (Section~\ref{3.1}) between GNN embeddings and PCA-processed LLM embeddings. 
Some methods have developed more complex fine-tuning modules based on GNNs. 
Pan {\it et al.}~\cite{DistillLLM} use a pair of GNNs for two-stage tuning. It first trains a GNN ``interpreter'' supervised by LLM-extracted keyword information, and then distills the interpreter to another downstream GNN. 
ENGINE~\cite{ENGINE} develops a bypass fine-tuning architecture for LLMs, ``G-Ladders'', which consists of multiple levels of GNNs and projectors. This structure enhances computational efficiency through node embedding caching.

Unlike GNNs, non-GNN fine-tuning module types such as MLPs lack the ability to handle local graph structures. To achieve modality alignment, these methods often resort to specialized tuning schemes. 
GraphGPT~\cite{GraphGPT} tunes a linear adapter through ``{\it graph-instruction matching}'' before downstream instruction tuning. The LLM is instructed to reorder the list of node text to match textual embeddings obtained from a parallel GNN+Transformer encoder.
LLaGA~\cite{LLaGA} extracts contextual knowledge from a graph using two tokenizers: node embedding concatenation through level-order traversal on a neighborhood tree, and neighborhood embedding aggregation in different hops. 
HIGHT~\cite{HIGHT} develops a hierarchical graph tokenizer to combine node, motif, and graph-level information for instruction tuning. 
GraphTranslator~\cite{GraphTranslator} tunes an attention-based adapter, the ``translator'', to project graph embeddings into the LLM space. It is guided by LLM-generated text that describes various knowledge patterns such as summaries of nodes, neighborhoods, and other commonalities.
\\[5pt]
{\bf Interfaces: representations vs. augmentations.} 
We have discussed GLMs that leverage LLM-empowered representations for fine-tuning. 
LLMs can also generate augmented data components for fine-tuning modules. They include:

(1) {\it Text}: 
TAPE~\cite{TAPE} instructs an LLM to explain its decisions in classifying nodes, facilitating an in-depth understanding of node-level information. These explanations are then encoded by a smaller LM to enrich textual features for downstream GNN training.
Unlike TAPE which directly uses original adjacency matrices, SFGL~\cite{SFGL} generates scale-free graphs from the input data to obtain textual features more aligned with real-world edge distributions.
KEA~\cite{KEA} instructs an LLM to generate descriptions of terminologies across different fields.
LLM4Mol~\cite{LLM4Mol} uses ChatGPT to generate descriptions of chemical molecules, including functional groups and other properties, to fine-tune a small-scale RoBERTa.
TANS~\cite{TANS} leverages centralities and clustering coefficients (Section~\ref{3.2}), along with contextual information, to generate descriptive text for nodes using GPT-4o-mini for non-textual graphs.
GAugLLM~\cite{GAugLLM} enhances textual representations with node context summaries, and uses them to guide feature-level and edge-level augmentations in self-supervised fine-tuning tasks such as node instance discrimination and masked feature prediction.

(2) {\it Pseudo-labels}:
LLM-GNN~\cite{LLM-GNN} leverages the LLM to generate cluster-aware node pseudo-labels to supervise a downstream GNN, referred to as {\it label-free node classification}. 
A set of nodes closer to $K$-means cluster centers is selected for annotation based on a cluster density metric. 
Similarly, Locle~\cite{Locle} uses subspace clustering to find the node set. LLM pseudo-labels are then selected based on their information certainty and refined by graph rewiring. 

(3) {\it Graph structure}:
LLM4NG~\cite{LLM4NG} and OpenGraph~\cite{OpenGraph} both use a pre-trained LLM to generate node samples and associated links to augment the original graph data. 
While LLM4NG trains an MLP-based edge predictor via cross-entropy-based link prediction, OpenGraph generates new edges through Gibbs sampling and trains a GT by masked link prediction.
Sun {\it et al.}~\cite{A-D+LPA} utilize GPT-3.5-Turbo to assist in removing unreliable edges and adding reliable ones to create a refined graph for GNN input. 
LOGIN~\cite{LOGIN} treats the LLM as a consultant for node classification during GNN fine-tuning. If the LLM prediction matches the ground truth, it updates the original feature; if not, it prunes its associated links based on neighbor similarity. 
\\[5pt]
{\bf Backbone parameters: frozen vs. tuned.} 
We have discussed GLMs with additional fine-tuning modules, where the LLM backbone remains frozen. 
Another line of studies designs special textual and graph domain co-training strategies to fine-tune the LLM backbone. 
GLEM~\cite{GLEM} iteratively tunes a GNN and a DeBERTa model using a variational Expectation-Maximization framework. 
GraphLLM~\cite{GraphLLM} synergistically tunes a GT and a LLaMA 2 model via ``prefix-tuning'', i.e., to project graph representations into trainable tokens and prepend them to the keys and values of every Transformer attention layer.
LEADING~\cite{LEADING} reduces the cost of joint LLM-GNN fine-tuning by decoupling the computation of node embedding from neighborhood embeddings.
Instead of joint tuning, SimTeG~\cite{SimTeG} separates the tuning of LMs and GNNs in a two-stage manner for more distinguishable embedding spaces.

Among all types of graph structural knowledge, link information (Section~\ref{3.3}) is often explicitly extracted to fine-tune LLMs, as it directly indicates relationships between entities and can be easily extracted by both GNNs and LLMs. 
InstructGLM~\cite{InstructGLM} introduces link prediction into LLM instruction tuning as an auxiliary objective, similar to the training of SuperGAT~\cite{SuperGAT}. 
CMRP~\cite{CMRP} tunes an LLM and a GNN to dynamically select an optimal edge set. The generated edge set is then injected into prompt tokens to iteratively instruct the LLM.
GIANT~\cite{GIANT} presents {\it neighborhood prediction} to fine-tune an XR-Transformer, which constructs hierarchical clusters and predicts the rows of the adjacency matrix as a multi-class classification task. 
AuGLM~\cite{AuGLM} selects and summarizes neighboring nodes by Personalized PageRank scoring and GNN-generated class prototypes for instruction tuning. 
Another type of knowledge information involves long-range paths (Section~\ref{4.2}). 
\conditionalblue{
WalkLM~\cite{WalkLM} incorporates long-range information into MLM to fine-tune a DistilRoBERTa model: it first samples attributed random walk sequences and then textualizes them as a token list.}
GUNDAM~\cite{GUNDAM} fine-tunes an LLM 
by reasoning different paths between two nodes, with path labels generated by unsupervised algorithms. 
LinguGKD~\cite{LinguGKD} develops a distillation architecture for LLM-GNNs: it first leverages node degrees and k-hop neighbors to instruct-tune an LLM. Then, the GNN is fine-tuned by contrasting every intermediate layer of both models.
InstructGraph~\cite{InstructGraph}, GraphWiz~\cite{GraphWiz}, and GraphInstruct~\cite{GraphInstruct} conduct multi-task LLM fine-tuning by combining over a dozen instruction tuning tasks. They include self-supervised predictions on various graph structures (cycles, shortest paths, maximum flows, Hamilton paths, etc.) as well as node-level and link-level downstream signals, aiming to provide the LLM with a comprehensive understanding of the graph domain. 
\\[5pt]
{\bf Discussion.}
While both GLM pre-training and tuning strategies demonstrate the potential of bridging the modality gap between graph and text, 
it remains underexplored whether GLMs have adequately tapped the potential in LLMs with billion-scale parameters. 
Some excellent properties of LLMs, e.g., the emergent ability~\cite{emergent}, are yet to be discovered on graph model architectures. 
The fragile side of LLMs such as 
hallucinations~\cite{hallucination} and intervention of spurious factors~\cite{NLGraph} keeps posing challenges to GLM researchers. 
To tackle these challenges, future research should focus on developing more powerful graph-specific architectures as well as pre-training and fine-tuning strategies that effectively extract and leverage various types of graph knowledge.

\section{Challenges and Future Directions} \label{8}

This section discusses potential challenges that graph researchers may encounter and insights for future research directions towards GFMs, as a conclusion to our survey.

\subsection{Combining Different Graph Knowledge Patterns} \label{8.1}

Despite that a variety of pretext tasks have been proposed for self-supervised graph models, 
the effectiveness of these pretexts depends on the application scenarios applied to downstream tasks~\cite{AutoSSL}.
Therefore, it is crucial to investigate how to effectively combine different graph knowledge to adapt graph models to more application scenarios.

One research direction is to jointly optimize different pretext objectives to obtain different aspects of graph knowledge. This can be formulated as a {\it multi-task pre-training} problem. 
Traditional methods simply assign hyperparameters to weigh each pretext task, leading to suboptimal performance. 
AutoSSL~\cite{AutoSSL} and ParetoGNN~\cite{ParetoGNN} are neural parameter search algorithms in order to dynamically find a set of optimal coefficients to combine different pretexts.
GraphTCM~\cite{GraphTCM} models a correlation value for every pair of pretexts and optimizes the correlation matrix to find the best parameters.
AGSSL~\cite{AGSSL} and WAS~\cite{WAS} propose a knowledge distillation technique, where the knowledge from different teachers is distilled into a single unified student.
Another promising approach is to design specific {\it multi-task prompting} strategies. 
ULTRA-DP~\cite{ULTRA-DP} and MultiGPrompt~\cite{MultiGPrompt} are multi-task prompting methods that assign a learnable task prompt for each pretext and pass them to the downstream model. 
While ULTRA-DP searches for the best task prompt for the downstream task, MultiGPrompt combines all of them by a linear combination or a parameterized network. 

\conditionalblue{
\subsection{Knowledge Adaptation across Graph Types} \label{8.2}

It is important for future GFMs to explore a wider range of graph knowledge, handling various complex data types beyond simple graphs. 
Here, we focus on several graph types and their corresponding knowledge patterns. We will show that these patterns and processing methods share commonalities with the graph knowledge discussed in Sections~\ref{3}--\ref{5}, implying potential data unification in future GFMs.
\\[5pt]
{\bf Heterogeneous \& knowledge graphs.}
Nodes and links in heterogeneous graphs possess unique types that exhibit distinct knowledge patterns. 
Some existing methods distinguish and handle different node and edge types. 
For instance, Heterformer~\cite{Heterformer} and RMR~\cite{RMR} are node instance discrimination methods within specific node and relation types, respectively. 
HiGPT~\cite{HiGPT} is a GLM that pre-trains an LLM by matching node types within sampled heterogeneous subgraphs.
For downstream tuning, HG-Adapter~\cite{HG-Adapter} is a PEFT approach that constructs a heterogeneous structure by calculating correlation scores for every neighboring node type to fine-tune an adapter. HGPROMPT~\cite{HGPROMPT} splits a heterogeneous graph into multiple type-specific subgraphs and designs a learnable prompt vector for each node type. 

However, these methods do not explicitly mine the higher-order structured knowledge underlying heterogeneous types.
For example, {\it meta-paths} are sequences of nodes and links with specific types, representing composite relations between different entities, such as ``co-write: author-paper-author'' and ``chemical reaction: compound-reaction-compound''. 
Learning meta-paths is similar to learning links, involving methods like meta-path prediction~\cite{SELAR,HGMAE,HierPromptLM} and instance discrimination with meta-path-based augmentation~\cite{HeCo}. Another example is HetGPT~\cite{HetGPT}, a prompting method that aggregates class prompts along meta-paths.
Additionally, the {\it network schema} is a motif-like heterogeneous pattern that contains all relationships of a particular node type. Due to the rich local structural information in the network schema and the ease of instance retrieval and extraction, it is used to generate contrastive instances in PT-HGNN~\cite{PT-HGNN} and HeCo~\cite{HeCo}.

Knowledge graphs are heterogeneous graphs embedded in specific domains and enriched with domain-specific factual knowledge. 
In knowledge graphs, {\it relation triples} (head entity, relation, tail entity) serve as the fundamental knowledge units. 
RotatE~\cite{RotatE} employs a negative sampling-based margin loss to discriminate between the head and tail entities in each relation triple. 
KEPLER~\cite{KEPLER} further incorporates the margin loss into MLM to fine-tune a RoBERTa model. 
SelfKG~\cite{SelfKG} and AutoAlign~\cite{AutoAlign} utilize an InfoNCE and Triplet estimator, respectively, to capture entity information and align them. 
GNP~\cite{GNP} proposes masked relation prediction, similar to masked link prediction, within constructed positive and negative relation triples.
\\[5pt]
{\bf Dynamic graphs.}
Real-world graph data often exhibits dynamic characteristics. 
The structure or size of a graph may change over time, a phenomenon known as {\it dynamic evolution}.
Discrete dynamic evolution manifests as a series of graph snapshots that are subgraphs with evolved features or graph structure, 
while continuous dynamic evolution is represented by a sequence of events with timestamps. 

Dynamic node features are coupled with temporal knowledge. 
GPT-ST~\cite{GPT-ST} and STGP~\cite{STGP} perform masked feature prediction across the time dimension of the feature matrix. DDGCL~\cite{DDGCL} introduces a temporal similarity function to the InfoNCE estimator, weighted by a time gap penalty. STGP~\cite{STGP} and DyGPrompt~\cite{DyGPrompt} design dual prompts that capture both downstream node features and time information. LLM4DyG~\cite{LLM4DyG} prompts an LLM to sequentially consider node and time information, significantly improving the LLM's reasoning abilities on dynamic graphs. 
For dynamic evolution on graph structures, GraphPro~\cite{GraphPro} constructs a graph prompt that concatenates various graph snapshots with the original structure, and incorporates temporal weights into downstream message passing.
\\[5pt]
{\bf Hypergraphs.}
Hypergraphs involve {\it hyperedges} that connect more than two nodes. They are better suitable for capturing higher-order structural relationships beyond simple pairwise interactions. 
Hyperedges can be considered as a special type of context, as they represent local commonalities among nodes.
VilLain~\cite{VilLain} learns a balanced and distinctive pseudo-label distribution by propagating the prototype vectors along the hyperedges. 
For instance discrimination models, HyperGCL~\cite{HyperGCL} focuses on generating hypergraph augmentations with a variational autoencoder, while TriCL~\cite{TriCL} focuses on performing contrast within and between nodes and hyperedges based on the connection membership. 
HypeBoy~\cite{HypeBoy} predicts if a node belongs to a hyperedge formed by another set of nodes by minimizing the similarity between their projected embeddings. 

\subsection{Tackling Potential Biases in Future GFMs} \label{8.3}

Potential biases in graph representations can significantly affect the generalization ability of graph models~\cite{ImGCL,RES}, presenting challenges in building GFMs. 
These biases may be inherent in the graph structural data itself or introduced during pre-training or downstream tuning, mainly including the following aspects.
\\[5pt]
{\bf Imbalanced data distributions.} 
Potential biases manifest when the feature or structural properties of graphs exhibit uneven distributions. 
They mainly include imbalanced class distributions, node degree distributions, and sensitive representation distributions. 
To mitigate potential biases, current methods involve constructing balanced training data distributions through structure-aware augmentations. 
GRADE~\cite{GRADE} capitalizes on feature similarity to balance nodes with imbalanced degrees, including removing dissimilar connections of high-degree nodes and aligning the neighbor distribution of similar low-degree nodes. 
For long-tailed class distributions, ImGCL~\cite{ImGCL} first divides a graph into several K-means clusters, each one referring to a latent class. 
Node representations are then sampled within each cluster based on PageRank scores. 
FPrompt~\cite{FPrompt} is a fair graph prompting approach where the structural prompt is obtained by masking links within different sensitive groups. 

Another way is to design specific training strategies to distinguish or emphasize imbalanced instances. 
CM-GCL~\cite{CM-GCL} prunes model parameters during contrastive pre-training to capture minority samples and employs a focal loss to emphasize these samples during fine-tuning. 
Graphair~\cite{Graphair} introduces a learnable graph augmentation network and a discriminative network to identify sensitive attributes in the augmented graphs. It utilizes adversarial training to update both networks. 
GraphPAR~\cite{GraphPAR} is a fair PEFT (Section~\ref{6.1}) approach that generates multiple sensitive attribute vectors for every node and minimizes the distance between them during fine-tuning.
\\[5pt]
{\bf Vulnerability to attacks.} 
Graph models that are not robust against perturbations are vulnerable to malicious attacks, 
such as injecting noise into features or textual attributes, and inserting nodes and links into the graph structure to disrupt the underlying knowledge patterns~\cite{GCBA,CrossBA}. 
Structural attacks often become more effective than feature-based ones~\cite{RES}.

To mitigate the impact of structural attacks, 
GRV~\cite{GRV} introduces a robustness quantification metric based on mutual information and uses it to guide the training of a DGI model~\cite{DGI}. 
RES~\cite{RES} demonstrates the effectiveness of random edge dropping in defense against structural attacks and applies it to node-level and graph-level instance discrimination. 
The recent emergence of GLMs has prompted further research in terms of their robustness. 
Although LLMs provide a certain degree of robustness compared to GNNs, their performance can still decline considerably under structural attacks~\cite{DeepDiveRobust,LLM4RGNN}. 
To improve structural robustness, LLM4RGNN~\cite{LLM4RGNN} tunes a local LLM and a small LM, which are used to identify and remove malicious links, as well as to re-add missing important links in the perturbed graph data. 
} 

\subsection{Building Powerful and Explainable GFMs} \label{8.4}

There is a pressing need for better explainability of GFMs. 
It not only ensures the model reliability in practical applications but also provides valuable insights into understanding graph knowledge. 
A promising and underexplored avenue is Reasoning on Graphs (RoG). 
Drawing from the huge success of LLM reasoning, RoG researchers tend to employ natural language instructions, such as Chain-of-Thoughts and Tree-of-Thoughts, to guide GLMs in reasoning over graph structures~\cite{GraphLLM,Graph-CoT,RoG}. 
Reasoning not only enhances the model explainability by reducing hallucinations but also enriches the knowledge provided, thereby boosting the generalization ability of GFMs in various open-ended tasks.
Additionally, recent GLMs leverage the power of external knowledge bases through Retrieval-Augmented Generation (RAG)~\cite{GRAGsurvey,G-Retriever,RAGraph}. 
These models retrieve knowledge relevant to the queries from knowledge graph databases to enhance reasoning explainability.
It is noteworthy that recent RAG approaches have begun to recognize the importance of higher-order structural knowledge, such as long-range paths~\cite{PathRAG} and clusters~\cite{ArchRAG}. They can be integrated into our knowledge taxonomy in the future.
By advancing in these directions, we can pave the way for versatile GFMs and graph agents that are capable of handling diverse, complex, and domain-specific graph tasks with unprecedented effectiveness and adaptability.

\bibliography{ref.bib}{}
\bibliographystyle{IEEEtran}


\onecolumn
\appendices

\conditionalblue{
\section{Summary} \label{A}

We have listed all references in our survey in Tables~\ref{pttable}--\ref{glmtable}. 
Table~\ref{pttable} summarizes self-supervised graph pre-training methods,
Table~\ref{dstable} summarizes graph downstream tuning methods,
and Table~\ref{glmtable} summarizes GLMs. 
Papers are arranged in strict chronological order determined by their earliest publication or preprinting time, indicated by the ``Time'' column.
For more information such as paper links and open-source code links, please refer to our GitHub list.
}

\begin{table*}[!h]
\caption{
\conditionalblue{
Summary of self-supervised graph pre-training methods.
}
}\label{pttable}
\centering
\adjustbox{max width=1.0\textwidth}{
\begin{tabular}{p{2.6cm}lp{1.5cm}p{5cm}p{3cm}p{3.8cm}}
\toprule[1.2pt]
Model & Time & Venue & Pre-training tasks & \makecell[l]{Graph knowledge\\focused on} & Downstream tasks
\\ \cmidrule(){1-6}
GAE; VGAE~\cite{GAE/VGAE} & Nov 2016 & \makecell[l]{NIPS\\Workshop\\(BDL)'16} & Link prediction & Links & Link prediction \\\rowcolor{gray!10}
GraphSAGE~\cite{GraphSAGE} & Jun 2017 & NIPS'17 & Context discrimination & Context & Node classification \\
MGAE~\cite{MGAE} & Nov 2017 & CIKM'17 & Feature prediction & Node features & Graph partitioning \\\rowcolor{gray!10}
ARGA; ARVGA~\cite{ARGA/ARVGA} & Feb 2018 & IJCAI'18 & Link prediction & Links & \makecell[l]{Link prediction;\\node clustering} \\
DGI~\cite{DGI} & Sept 2018 & ICLR'19 & Node-graph discrimination & Global structure & Node classification \\\rowcolor{gray!10}
KernelPred~\cite{KernelPT} & Nov 2018 & arXiv & Graph similarity prediction & Global structure & Graph classification \\
RotatE~\cite{RotatE} & Feb 2019 & ICLR'19 & Node instance discrimination & Node features; links & Link prediction \\\rowcolor{gray!10}
M3S~\cite{M3S} & Feb 2019 & AAAI'20 & Node clustering & Clusters & Node classification \\
\makecell[l]{Hu {\it et al.}~\cite{UPGCN}\\(ScoreRank;\\DenoisingRecon;\\ClusterDetect)} & May 2019 & \makecell[l]{ICLR\\Workshop\\(RLGM)'19} & \makecell[l]{Centrality ranking\\/ masked link prediction\\/ graph partitioning} & \makecell[l]{Node properties; links;\\clusters} & Node classification \\\rowcolor{gray!10}
\makecell[l]{GNN-Pretrain~\cite{GNNpretrain}\\(AttrMask;\\ContextPred)} & May 2019 & ICLR'20 & \makecell[l]{Masked feature prediction\\/ edge feature prediction\\/ contextual subgraph discrimination}& \makecell[l]{Node features; links;\\context} & \makecell[l]{Graph classification;\\biological function prediction} \\
InfoGraph~\cite{InfoGraph} & Jul 2019 & ICLR'20 & Node-graph discrimination & \makecell[l]{Context;\\global structure} & Graph classification \\\rowcolor{gray!10}
GALA~\cite{GALA} & Aug 2019 & ICCV'19 & Feature prediction & Node features & \makecell[l]{Node clustering;\\link prediction; etc.} \\
SIG-VAE~\cite{SIG-VAE} & Aug 2019 & NeurIPS'19 & Link prediction & Links & \makecell[l]{Node classification;\\link prediction;\\node clustering;\\graph generation} \\\rowcolor{gray!10}
Graph-Bert~\cite{Graph-Bert} & Jan 2020 & arXiv & \makecell[l]{Feature prediction;\\similarity prediction} & \makecell[l]{Node features; context;\\long-range similarities} & \makecell[l]{Node classification;\\node clustering} \\
GMI~\cite{GMI} & Feb 2020 & WWW'20 & Feature prediction & Node features & \makecell[l]{Node classification;\\link prediction} \\\rowcolor{gray!10}
S\textsuperscript{2}GRL~\cite{S2GRL} & Mar 2020 & \makecell[l]{Information\\Sciences'22} & Similarity prediction & Long-range similarities & \makecell[l]{Node classification;\\node clustering;\\link prediction} \\
GRACE~\cite{GRACE} & Jun 2020 & \makecell[l]{ICML\\Workshop\\(GRL+)'20} & Node instance discrimination & Node features & Node classification \\\rowcolor{gray!10}
MVGRL~\cite{MVGRL} & Jun 2020 & ICML'20 & Node-graph discrimination & Global structure & \makecell[l]{Node classification;\\graph classification} \\
\makecell[l]{SS-GCN~\cite{SS-GCN}\\(GraphComp;\\NodeCluster;\\GraphPar)} & Jun 2020 & ICML'20 & \makecell[l]{Masked feature prediction\\/ node clustering\\/ graph partitioning} & Node features; clusters & Node classification \\\rowcolor{gray!10}
GCC~\cite{GCC} & Jun 2020 & KDD'20 & Contextual subgraph discrimination & Context & \makecell[l]{Node classification;\\graph classification;\\similarity search} \\
\makecell[l]{SelfTask~\cite{SelfTask}\\(AttributeMask;\\NodeProperty;\\EdgeMask;\\PairwiseDistance;\\PairwiseAttrSim;\\Distance2Clusters)} & Jun 2020 & arXiv & \makecell[l]{Masked feature prediction\\/ property prediction\\/ masked link prediction\\/ similarity prediction\\/ graph partitioning} & \makecell[l]{Node features;\\node properties; links;\\long-range similarities;\\clusters} & Node classification \\\rowcolor{gray!10}
GROVER~\cite{GROVER} & Jun 2020 & NeurIPS'20 & \makecell[l]{Motif prediction;\\contextual property prediction} & Context; motifs & \makecell[l]{Graph classification;\\graph regression} \\
GPT-GNN~\cite{GPT-GNN} & Jun 2020 & KDD'20 & \makecell[l]{Masked feature prediction;\\masked link prediction} & Node features; links & \makecell[l]{Node classification;\\link prediction;\\edge classification} \\\rowcolor{gray!10}
AGE~\cite{AGE} & Jul 2020 & KDD'20 & Similarity prediction & Long-range similarities & \makecell[l]{Node clustering;\\link prediction} \\
AM-GCN~\cite{AM-GCN} & Jul 2020 & KDD'20 & Similarity graph alignment & Long-range similarities & Node classification \\\rowcolor{gray!10}
SELAR~\cite{SELAR} & Jul 2020 & NeurIPS'20 & Link prediction & Links & \makecell[l]{(Heterogeneous)\\node classification;\\link prediction} \\
EGI~\cite{EGI} & Sept 2020 & NeurIPS'21 & Context discrimination & Context & \makecell[l]{Link prediction;\\structural role identification} \\\rowcolor{gray!10}
Subg-Con~\cite{Subg-Con} & Sept 2020 & ICDM'20 & Context discrimination & Context & Node classification \\
\bottomrule[1.2pt]
\end{tabular}
}
\end{table*}

\setcounter{table}{0}
\begin{table*}[!t]
\caption{
\conditionalblue{
Summary of self-supervised graph pre-training methods (continued).
}
}
\centering
\adjustbox{max width=1.0\textwidth}{
\begin{tabular}{p{2.6cm}lp{1.5cm}p{5cm}p{3cm}p{3.8cm}}
\toprule[1.2pt]
Model & Time & Venue & Pre-training tasks & \makecell[l]{Graph knowledge\\focused on} & Downstream tasks
\\ \cmidrule(){1-6}
CG\textsuperscript{3}~\cite{CG3} & Sept 2020 & AAAI'21 & \makecell[l]{Node instance discrimination;\\link prediction} & Node features; links & Node classification \\\rowcolor{gray!10}
DGVAE~\cite{DGVAE} & Oct 2020 & NeurIPS'20 & Partition-conditioned link prediction & Links; clusters & \makecell[l]{Node clustering;\\graph generation} \\
CommDGI~\cite{CommDGI} & Oct 2020 & CIKM'20 & \makecell[l]{Cluster-based discrimination;\\graph partitioning} & Clusters & Node clustering \\\rowcolor{gray!10}
GraphCL~\cite{GraphCL} & Oct 2020 & NeurIPS'20 & Graph instance discrimination & Global structure & Graph classification \\
GCA~\cite{GCA} & Oct 2020 & WWW'21 & Node instance discrimination & Node features & Node classification \\\rowcolor{gray!10}
GRV~\cite{GRV} & Dec 2020 & AAAI'22 & Node-graph discrimination & Global structure & \makecell[l]{Node classification;\\node clustering;\\link prediction} \\
MICRO-Graph~\cite{MICRO-Graph} & Dec 2020 & TKDE'24 & \makecell[l]{Motif-based discrimination;\\graph partitioning} & Motifs; clusters & Graph classification \\
Mask-GVAE~\cite{Mask-GVAE} & Feb 2021 & WWW'21 & \makecell[l]{Graph partitioning;\\partition-conditioned link prediction} & Clusters & Node clustering; etc. \\\rowcolor{gray!10}
SLAPS~\cite{SLAPS} & Feb 2021 & NeurIPS'21 & Masked feature prediction & \makecell[l]{Node features;\\long-range similarities} & Node classification; etc. \\
BGRL~\cite{BGRL} & Feb 2021 & ICLR'22 & Node instance discrimination & Node features & Node classification \\\rowcolor{gray!10}
PIGAE~\cite{PIGAE} & Apr 2021 & NeurIPS'21 & \makecell[l]{Node order matching;\\link prediction;\\edge feature prediction} & Node properties; links & Graph classification \\
VICReg~\cite{VICReg} & May 2021 & ICLR'22 & \makecell[l]{Node instance discrimination;\\dimension discrimination} & Node features & Node classification; etc. \\\rowcolor{gray!10}
MERIT~\cite{MERIT} & May 2021 & IJCAI'21 & Node instance discrimination & Node features & Node classification \\
MVMI-FT~\cite{MVMI-FT} & May 2021 & TKDE'23 & \makecell[l]{Link prediction;\\node-graph discrimination;\\similarity graph alignment} & \makecell[l]{Links;\\long-range similarities;\\global structure} & \makecell[l]{Node classification;\\node clustering} \\\rowcolor{gray!10}
HeCo~\cite{HeCo} & May 2021 & KDD'21 & Node instance discrimination & Node features; motifs & \makecell[l]{(Heterogeneous)\\node classification;\\node clustering} \\
G-BT~\cite{G-BT} & Jun 2021 & KBS'22 & Dimension discrimination & Node features & Node classification \\\rowcolor{gray!10}
Graph-MLP~\cite{Graph-MLP} & Jun 2021 & arXiv & Context discrimination & Context & Node classification \\
GraphLoG~\cite{GraphLoG} & Jun 2021 & ICML'21 & \makecell[l]{Contextual subgraph discrimination;\\node clustering;\\graph instance discrimination} & \makecell[l]{Context; clusters;\\global structure} & \makecell[l]{Graph classification;\\biological function prediction} \\\rowcolor{gray!10}
AutoSSL~\cite{AutoSSL} & Jun 2021 & ICLR'22 & Miscellaneous & -- & \makecell[l]{Node classification;\\node clustering} \\
JOAO~\cite{JOAO} & Jun 2021 & ICML'21 & Graph instance discrimination & Global structure & Graph classification \\\rowcolor{gray!10}
AD-GCL~\cite{AD-GCL} & Jun 2021 & NeurIPS'21 & Graph instance discrimination & Global structure & Graph classification \\
CCA-SSG~\cite{CCA-SSG} & Jun 2021 & NeurIPS'21 & \makecell[l]{Node instance discrimination;\\dimension discrimination} & Node features & Node classification \\\rowcolor{gray!10}
PT-HGNN~\cite{PT-HGNN} & Aug 2021 & KDD'21 & Context discrimination & \makecell[l]{Node features; context;\\motifs} & \makecell[l]{(Heterogeneous)\\node classification;\\link prediction} \\
MGSSL~\cite{MGSSL} & Oct 2021 & NeurIPS'21 & \makecell[l]{Masked feature prediction;\\masked edge feature prediction;\\motif prediction} & \makecell[l]{Node features; links;\\motifs} & Graph classification \\\rowcolor{gray!10}
ProGCL~\cite{ProGCL} & Oct 2021 & ICML'22 & Node instance discrimination & Node features & Node classification \\
gCooL~\cite{gCooL} & Oct 2021 & WWW'22 & Partition-based discrimination & Clusters & \makecell[l]{Node classification;\\node clustering} \\\rowcolor{gray!10}
DDGCL~\cite{DDGCL} & Oct 2021 & CIKM'21 & Node instance discrimination & Node features & \makecell[l]{(Temporal)\\node classification;\\link prediction} \\
AFGRL~\cite{AFGRL} & Dec 2021 & AAAI'22 & Context discrimination & Node features; context & \makecell[l]{Node classification;\\node clustering;\\similarity search} \\\rowcolor{gray!10}
SelfMGNN~\cite{SelfMGNN} & Dec 2021 & AAAI'22 & Cross-manifold discrimination & Manifolds & Node classification \\
SimGCL~\cite{SimGCL} & Dec 2021 & SIGIR'22 & Node instance discrimination & Node features & Recommendation \\\rowcolor{gray!10}
S2GAE~\cite{S2GAE} & Jan 2022 & WSDM'23 & Masked link prediction & Links & \makecell[l]{Node classification;\\graph classification;\\link prediction} \\
COLES~\cite{COLES} & Jan 2022 & NeurIPS'21 & Context discrimination & Context & \makecell[l]{Node classification;\\node clustering} \\\rowcolor{gray!10}
DSGC~\cite{DSGC} & Jan 2022 & WWW'22 & Cross-manifold discrimination & Manifolds & Graph classification \\
HGCL~\cite{HGCL} & Jan 2022 & \makecell[l]{NeurIPS\\Workshop\\(SSL)'21} & Cross-manifold discrimination & Manifolds & Node classification \\\rowcolor{gray!10}
D-SLA~\cite{D-SLA} & Feb 2022 & NeurIPS'22 & \makecell[l]{Group discrimination;\\graph similarity prediction} & Global structure & \makecell[l]{Graph classification;\\link prediction} \\
SimGRACE~\cite{SimGRACE} & Feb 2022 & WWW'22 & Graph instance discrimination & Global structure & Graph classification \\\rowcolor{gray!10}
LaGraph~\cite{LaGraph} & Feb 2022 & ICML'22 & \makecell[l]{Masked feature prediction;\\node instance discrimination} & Node features; context & \makecell[l]{Node classification;\\graph classification} \\
S\textsuperscript{3}-CL~\cite{S3-CL} & Feb 2022 & AAAI'23 & \makecell[l]{Contextual subgraph discrimination;\\cluster-based discrimination} & Context; clusters & \makecell[l]{Node classification;\\node clustering} \\\rowcolor{gray!10}
NWR-GAE~\cite{NWR-GAE} & Feb 2022 & ICLR'22 & \makecell[l]{Property prediction;\\context feature prediction} & \makecell[l]{Node properties;\\context} & \makecell[l]{Node classification;\\structural role identification} \\
N2N~\cite{N2N} & Mar 2022 & CVPR'22 & Context discrimination & Context & Node classification \\
\bottomrule[1.2pt]
\end{tabular}
}
\end{table*}

\setcounter{table}{0}
\begin{table*}[!t]
\caption{
\conditionalblue{
Summary of self-supervised graph pre-training methods (continued).
}
}
\centering
\adjustbox{max width=1.0\textwidth}{
\begin{tabular}{p{2.6cm}lp{1.5cm}p{5cm}p{3cm}p{3.8cm}}
\toprule[1.2pt]
Model & Time & Venue & Pre-training tasks & \makecell[l]{Graph knowledge\\focused on} & Downstream tasks
\\ \cmidrule(){1-6}
\rowcolor{gray!10}
SuperGAT~\cite{SuperGAT} & Apr  2022 & ICLR'21 & Link prediction & Links & \makecell[l]{Node classification;\\link prediction} \\
MaskGAE~\cite{MaskGAE} & May 2022 & KDD'23 & \makecell[l]{Property prediction;\\masked link prediction} & \makecell[l]{Node properties; links;\\long-range similarities} & \makecell[l]{Node classification;\\link prediction} \\\rowcolor{gray!10}
Heterformer~\cite{Heterformer} & May 2022 & KDD'23 & Node instance discrimination & Node features & \makecell[l]{(Heterogeneous)\\node classification;\\node clustering;\\link prediction} \\
GraphMAE~\cite{GraphMAE} & May 2022 & KDD'22 & Masked feature prediction & Node features & \makecell[l]{Node classification;\\graph classification} \\\rowcolor{gray!10}
ImGCL~\cite{ImGCL} & May 2022 & AAAI'23 & Node instance discrimination & \makecell[l]{Node features;\\node properties;\\clusters} & Node classification \\
GGD~\cite{GGD} & Jun 2022 & NeurIPS'22 & Group discrimination & Global structure & Node classification \\\rowcolor{gray!10}
COSTA~\cite{COSTA} & Jun 2022 & KDD'22 & Node instance discrimination & Node features & Node classification \\
TriCL~\cite{TriCL} & Jun 2022 & AAAI'23 & \makecell[l]{Node instance discrimination;\\context discrimination} & \makecell[l]{Node features; links;\\context} & \makecell[l]{Node classification;\\node clustering} \\\rowcolor{gray!10}
SUGRL~\cite{SUGRL} & Jun 2022 & AAAI'22 & \makecell[l]{Node instance discrimination;\\context discrimination} & Node features; context & Node classification \\
CGC~\cite{CGC} & Jul 2022 & WWW'23 & Graph instance discrimination & Global structure & Graph classification \\\rowcolor{gray!10}
HGMAE~\cite{HGMAE} & Aug 2022 & AAAI'23 & \makecell[l]{Feature prediction;\\masked feature prediction;\\masked link prediction} & Node features; links & \makecell[l]{(Heterogeneous)\\node classification;\\node clustering} \\
SPAN~\cite{SPAN} & Oct 2022 & ICLR'23 & Node-graph discrimination & \makecell[l]{Global structure;\\spectrum} & \makecell[l]{Node classification;\\graph classification;\\graph regression} \\\rowcolor{gray!10}
ParetoGNN~\cite{ParetoGNN} & Oct 2022 & ICLR'23 & Miscellaneous & -- & \makecell[l]{Node classification;\\node clustering;\\graph partitioning;\\link prediction} \\
AGSSL~\cite{AGSSL} & Oct 2022 & arXiv & Miscellaneous & -- & Node classification \\\rowcolor{gray!10}
GRADE~\cite{GRADE} & Oct 2022 & NeurIPS'22 & Node instance discrimination & \makecell[l]{Node features;\\node properties;\\long-range similarities} & Node classification \\
HyperGCL~\cite{HyperGCL} & Oct 2022 & NeurIPS'22 & \makecell[l]{Feature prediction;\\node instance discrimination;\\link prediction} & Node features; links & \makecell[l]{Node classification;\\(hyper-)link prediction} \\\rowcolor{gray!10}
HGRL~\cite{HGRL} & Oct 2022 & CIKM'22 & Context discrimination & Context & \makecell[l]{Node classification;\\node clustering} \\
CGI~\cite{CGI} & Nov 2022 & NeurIPS'22 & Node instance discrimination & Node features & Recommendation \\\rowcolor{gray!10}
GLEN~\cite{GLEN} & Nov 2022 & NeurIPS'22 & Context discrimination & Context & \makecell[l]{Node classification;\\node clustering} \\
CM-GCL~\cite{CM-GCL} & Nov 2022 & NeurIPS'22 & \makecell[l]{Node instance discrimination;\\similarity-based discrimination} & \makecell[l]{Node features;\\long-range similarities} & Node classification \\\rowcolor{gray!10}
SP-GCL~\cite{SP-GCL} & Nov 2022 & TMLR'23 & Node instance discrimination & Node features; context & Node classification \\
Mole-BERT~\cite{Mole-BERT} & Feb 2023 & ICLR'23 & \makecell[l]{Masked feature prediction;\\graph instance discrimination} & \makecell[l]{Node features;\\global structure} & \makecell[l]{Graph classification;\\graph regression} \\\rowcolor{gray!10}
Graphair~\cite{Graphair} & Feb 2023 & ICLR'23 & \makecell[l]{Masked feature prediction;\\masked link prediction;\\node instance discrimination} & Node features; links & Node classification \\
LightGCL~\cite{LightGCL} & Feb 2023 & ICLR'23 & Node instance discrimination & Node features & Recommendation \\\rowcolor{gray!10}
GraphMAE2~\cite{GraphMAE2} & Apr 2023 & WWW'23 & \makecell[l]{Masked feature prediction;\\node instance discrimination} & Node features & Node classification \\
CSGCL~\cite{CSGCL} & May 2023 & IJCAI'23 & Partition-based discrimination & Clusters & \makecell[l]{Node classification;\\node clustering;\\link prediction} \\\rowcolor{gray!10}
CARL-G~\cite{CARL-G} & Jun 2023 & KDD'23 & Node clustering & Clusters & \makecell[l]{Node classification;\\node clustering;\\similarity search} \\
HomoGCL~\cite{HomoGCL} & Jun 2023 & KDD'23 & \makecell[l]{Node clustering;\\cluster-based discrimination} & Clusters & \makecell[l]{Node classification;\\node clustering} \\\rowcolor{gray!10}
AEGCL~\cite{AEGCL} & Jun 2023 & TKDE'23 & \makecell[l]{Feature prediction;\\link prediction;\\similarity graph alignment} & \makecell[l]{Node features; links;\\long-range similarities} & \makecell[l]{Node classification;\\node clustering;\\link prediction} \\
DLR-GAE~\cite{DLR-GAE} & Jun 2023 & AAAI'23 & \makecell[l]{Link prediction;\\similarity graph alignment} & \makecell[l]{Links;\\long-range similarities} & Node classification \\\rowcolor{gray!10}
DGSI~\cite{DGSI} & Jun 2023 & AAAI'23 & \makecell[l]{Context discrimination;\\node-graph discrimination} & \makecell[l]{Context;\\global structure} & Node classification \\
ASP~\cite{ASP} & Jun 2023 & AAAI'23 & Similarity graph alignment & Long-range similarities & Node classification \\\rowcolor{gray!10}
MoAMa~\cite{MoAMa} & Sept 2023 & LoG'24 & Motif-based masked feature prediction & Node features; motifs & Graph classification \\
Graph-JEPA~\cite{Graph-JEPA} & Sept 2023 & TMLR'25 & Hyperbolic angle prediction & Context; manifolds & \makecell[l]{Graph classification;\\graph regression} \\\rowcolor{gray!10}
GraphFP~\cite{GraphFP} & Oct 2023 & NeurIPS'23 & \makecell[l]{Motif prediction;\\motif-based discrimination} & Motifs & \makecell[l]{Graph classification;\\graph regression} \\
RES~\cite{RES} & Oct 2023 & NeurIPS'23 & \makecell[l]{Node instance discrimination\\/ graph instance discrimination} & \makecell[l]{Node features;\\global structure} & \makecell[l]{Node classification\\/ graph classification} \\\rowcolor{gray!10}
GPT-ST~\cite{GPT-ST} & Nov 2023 & arXiv & Masked feature prediction & Node features; clusters & Time series forecasting \\
\bottomrule[1.2pt]
\end{tabular}
}
\end{table*}

\setcounter{table}{0}
\begin{table*}[!t]
\caption{
\conditionalblue{
Summary of self-supervised graph pre-training methods (continued).
}
}
\centering
\adjustbox{max width=1.0\textwidth}{
\begin{tabular}{p{2.6cm}lp{1.5cm}p{5cm}p{3cm}p{3.8cm}}
\toprule[1.2pt]
Model & Time & Venue & Pre-training tasks & \makecell[l]{Graph knowledge\\focused on} & Downstream tasks
\\ \cmidrule(){1-6}
StructComp~\cite{StructComp} & Dec 2023 & ICLR'24 & Partition-based discrimination & Clusters & Node classification \\\rowcolor{gray!10}
DGPM~\cite{DGPM} & Dec 2023 & AAAI'24 & \makecell[l]{Masked feature prediction;\\motif prediction} & Node features; motifs & Graph classification \\
HTML~\cite{HTML} & Dec 2023 & AAAI'24 & \makecell[l]{Contextual property prediction;\\graph instance discrimination;\\graph similarity prediction} & \makecell[l]{Context;\\global structure} & Graph classification \\\rowcolor{gray!10}
MotifRGC~\cite{MotifRGC} & Jan 2024 & AAAI'24 & \makecell[l]{Motif-based discrimination;\\cross-manifold discrimination} & Motifs; manifolds & \makecell[l]{Node classification;\\link prediction} \\
HypeBoy~\cite{HypeBoy} & Jan 2024 & ICLR'24 & Context discrimination & Node features; context & \makecell[l]{Node classification;\\(hyper-)link prediciton} \\\rowcolor{gray!10}
CTAug~\cite{CTAug} & Jan 2024 & WWW'24 & \makecell[l]{Node instance discrimination;\\motif-based discrimination} & Node features; motifs & Node classification \\
Bandana~\cite{Bandana} & Feb 2024 & WWW'24 & Link denoising & Links & \makecell[l]{Node classification;\\link prediction} \\\rowcolor{gray!10}
DCGL~\cite{DCGL} & Feb 2024 & AAAI'24 & \makecell[l]{Feature prediction;\\similarity graph alignment;\\cluster-based discrimination} & \makecell[l]{Node features;\\long-range similarities;\\clusters} & Node clustering \\
HASH-CODE~\cite{HASH-CODE} & Feb 2024 & WWW'24 & \makecell[l]{Node instance discrimination;\\context discrimination;\\contextual subgraph discrimination} & Node features; context & \makecell[l]{Node classification;\\link prediction} \\\rowcolor{gray!10}
WAS~\cite{WAS} & Mar 2024 & ICLR'24 & Miscellaneous & -- & \makecell[l]{Node classification;\\graph classification} \\
ASD-VAE~\cite{ASD-VAE} & Mar 2024 & WSDM'24 & \makecell[l]{Feature prediction;\\edge feature prediction} & Node features; links & Node classification; etc. \\\rowcolor{gray!10}
MGSE~\cite{MGSE} & May 2024 & ICML'24 & Node clustering & Clusters & Graph classification \\
GraphTCM~\cite{GraphTCM} & May 2024 & ICML'24 & Miscellaneous & -- & \makecell[l]{Node classification;\\link prediction} \\\rowcolor{gray!10}
D-VGAE~\cite{D-VGAE} & May 2024 & WWW'24 & Link prediction & Links; clusters & \makecell[l]{Node classification;\\node clustering;\\link prediction} \\
VilLain~\cite{VilLain} & May 2024 & WWW'24 & Node clustering & Links; clusters & \makecell[l]{Node classification;\\node clustering;\\(hyper-)link prediciton; etc.} \\\rowcolor{gray!10}
STGP~\cite{STGP} & May 2024 & CIKM'24 & Masked feature prediction & Node features & Time series forcasting \\
DiscoGNN~\cite{DiscoGNN} & Jul 2024 & ICDE'24 & \makecell[l]{Masked feature prediction;\\edge feature prediction;\\graph instance discrimination} & \makecell[l]{Node features;links;\\global structure} & \makecell[l]{Graph classification;\\similarity search} \\\rowcolor{gray!10}
LogDet~\cite{LogDet} & Aug 2024 & KDD'24 & Dimension discrimination & Node features & Node classification \\
RMR~\cite{RMR} & Aug 2024 & KDD'24 & Node instance discrimination & Node features; links & \makecell[l]{(Heterogeneous)\\node classification} \\\rowcolor{gray!10}
SGRL~\cite{SGRL} & Sept 2024 & NeurIPS'24 & Node instance discrimination & Node features; context & \makecell[l]{Node classification;\\node clustering} \\
HDM-GAE~\cite{HDM-GAE} & Jan 2025 & COLING'25 & Hyperbolic masked prediction & \makecell[l]{Node features; links;\\manifolds} & \makecell[l]{Node classification;\\link prediction} \\\rowcolor{gray!10}
CenPre~\cite{CenPre} & Jan 2025 & ICLR'25 & \makecell[l]{Node instance discrimination;\\property prediction} & \makecell[l]{Node features;\\node properties} & \makecell[l]{Node classification;\\graph classification;\\link prediction} \\
BSG~\cite{BSG} & Jan 2025 & WWW'25 & \makecell[l]{Node instance discrimination;\\context discrimination} & Node features; context & \makecell[l]{Node classification;\\link prediction} \\\rowcolor{gray!10}
RiemannGFM~\cite{RiemannGFM} & Jan 2025 & WWW'25 & Cross-manifold discrimination & Manifolds & \makecell[l]{Node classification;\\link prediction} \\
\bottomrule[1.2pt]
\end{tabular}
}
\end{table*}

\begin{table*}[!t]
\setlength{\tabcolsep}{7.5pt}
\caption{
\conditionalblue{
Summary of graph downstream tuning methods. ``SFT'' refers to ``supervised fine-tuning''.
}
}\label{dstable}
\centering
\adjustbox{max width=1.0\textwidth}{
\begin{tabular}{llp{1.5cm}p{1.6cm}p{5cm}p{3cm}p{3cm}}
\toprule[1.2pt]
Model & Time & Venue & \makecell[l]{Tuning\\strategy} & \trainicon Training/\tuneicon tuning tasks & \makecell[l]{Graph knowledge\\focused on} & Downstream tasks
\\ \cmidrule(){1-7}
L2P-GNN~\cite{L2P-GNN} & May 2021 & AAAI'21 & Fine-tuning & \makecell[l]{\trainicon Context discrimination; \\\hspace{7.5pt} graph instance discrimination \tuneicon SFT} & \makecell[l]{Node features; context;\\global structure} & Graph classification \\\rowcolor{gray!10}
AUX-TS~\cite{AUX-TS} & Jul 2021 & AAAI'21 & Fine-tuning & \makecell[l]{\trainicon Masked feature prediction; \\\hspace{7.5pt} masked link prediction \tuneicon SFT} & Node features; links & \makecell[l]{Node classification; \\link prediction }\\
GTOT-Tuning~\cite{GTOT} & Mar 2022 & IJCAI'22 & Fine-tuning & \trainicon Miscellaneous \tuneicon SFT & -- & Graph classification \\\rowcolor{gray!10}
GPPT~\cite{GPPT} & Aug 2022 & KDD'22 & Prompting & \makecell[l]{\trainicon Masked link prediction \tuneicon SFT} & Links; context; clusters & Node classification \\
GPF~\cite{GPF} & Sept 2022 & NeurIPS'23 & Prompting & \trainicon Miscellaneous \tuneicon SFT & Node features & \makecell[l]{Node classification; \\graph classification; \\link prediction }\\\rowcolor{gray!10}
GraphPrompt~\cite{GraphPrompt} & Feb 2023 & WWW'23 & Prompting & \makecell[l]{\trainicon Context discrimination \tuneicon SFT} & Context & \makecell[l]{Node classification; \\graph classification }\\
SGL-PT~\cite{SGL-PT} & Feb 2023 & arXiv & Prompting & \makecell[l]{\trainicon Masked feature prediction; \\\hspace{7.5pt} graph instance discrimination\\\tuneicon Masked feature prediction; SFT} & \makecell[l]{Node features; \\global structure} & \makecell[l]{Node classification; \\graph classification }\\\rowcolor{gray!10}
W2PGNN~\cite{W2PGNN} & Mar 2023 & KDD'23 & Fine-tuning & \trainicon Miscellaneous \tuneicon SFT & Motifs & \makecell[l]{Node classification; \\graph classification }\\
AdapterGNN~\cite{AdapterGNN} & Apr 2023 & AAAI'24 & \makecell[l]{Fine-tuning\\(PEFT)} & \trainicon Miscellaneous \tuneicon SFT & -- & Graph classification \\\rowcolor{gray!10}
G-Adapter~\cite{G-Adapter} & May 2023 & AAAI'24 & \makecell[l]{Fine-tuning\\(PEFT)} & \trainicon Miscellaneous \tuneicon SFT & \makecell[l]{Node features; \\long-range similarities} & Graph classification \\
PRODIGY~\cite{PRODIGY} & May 2023 & NeurIPS'23 & Prompting & \makecell[l]{\trainicon Miscellaneous\\\tuneicon Context discrimination; SFT} & Context & \makecell[l]{Node classification; \\graph classification; \\link prediction }\\\rowcolor{gray!10}
VNT~\cite{VNT} & Jun 2023 & KDD'23 & Prompting & \trainicon Miscellaneous \tuneicon SFT & Node features & \makecell[l]{Node classification; \\node clustering }\\
All in One~\cite{AllinOne} & Jul 2023 & KDD'23 & Prompting & \trainicon Miscellaneous \tuneicon SFT & Node features; links & \makecell[l]{Node classification; \\graph classification; \\link prediction; \\edge regression; \\graph regression }\\\rowcolor{gray!10}
S2PGNN~\cite{S2PGNN} & Aug 2023 & ICDE'24 & Fine-tuning & \trainicon Miscellaneous \tuneicon SFT & Global structure & \makecell[l]{Graph classification; \\graph regression }\\
DeepGPT~\cite{DeepGPT} & Sept 2023 & arXiv & Prompting & \trainicon Miscellaneous \tuneicon SFT & Node features & \makecell[l]{Graph classification; \\graph regression }\\\rowcolor{gray!10}
GraphControl~\cite{GraphControl} & Oct 2023 & WWW'24 & \makecell[l]{Fine-tuning;\\prompting} & \trainicon Miscellaneous \tuneicon SFT & Node features & Node classification \\
OFA~\cite{OFA} & Oct 2023 & ICLR'24 & Prompting & \tuneicon SFT & Node features; context & \makecell[l]{Node classification; \\graph classification; \\link prediction }\\\rowcolor{gray!10}
Self-Pro~\cite{Self-Pro} & Oct 2023 & \makecell[l]{ECML-\\PKDD'24} & Prompting & \makecell[l]{\trainicon Context discrimination \tuneicon SFT} & Node features; context & \makecell[l]{Node classification; \\link prediction }\\
ULTRA-DP~\cite{ULTRA-DP} & Oct 2023 & arXiv & Prompting & \makecell[l]{\trainicon Link prediction; \\\hspace{7.5pt} context discrimination \tuneicon SFT} & \makecell[l]{Node features; \\links; context} & \makecell[l]{Node classification; \\link prediction }\\\rowcolor{gray!10}
HetGPT~\cite{HetGPT} & Oct 2023 & WWW'24 & Prompting & \makecell[l]{\trainicon Node instance discrimination \tuneicon SFT} & \makecell[l]{Node features; \\links; context} & \makecell[l]{(Heterogeneous)\\node classification} \\
PSP~\cite{PSP} & Oct 2023 & \makecell[l]{ECML-\\PKDD'24} & Prompting & \makecell[l]{\trainicon Node-text discrimination \tuneicon SFT} & Node features & \makecell[l]{Node classification; \\graph classification }\\\rowcolor{gray!10}
GraphPrompt+~\cite{GraphPrompt+} & Nov 2023 & TKDE'24 & Prompting & \trainicon Miscellaneous \tuneicon SFT & \makecell[l]{Context; \\global structure} & \makecell[l]{Node classification; \\graph classification }\\
GraphPro~\cite{GraphPro} & Nov 2023 & WWW'24 & Prompting & \makecell[l]{\trainicon Node instance discrimination \\\tuneicon Node instance discrimination} & Node features; links & Recommendation\\\rowcolor{gray!10}
HGPROMPT~\cite{HGPROMPT} & Dec 2023 & AAAI'24 & Prompting & \makecell[l]{\trainicon Link prediction \tuneicon SFT} & \makecell[l]{Node features; \\links; context} & \makecell[l]{(Heterogeneous)\\node classification; \\graph classification }\\
MultiGPrompt~\cite{MultiGPrompt} & Dec 2023 & WWW'24 & Prompting & \makecell[l]{\trainicon Link prediction; \\\hspace{7.5pt} graph instance discrimination; \\\hspace{7.5pt} node-graph discrimination \tuneicon SFT} & Links; global structure & \makecell[l]{Node classification; \\graph classification }\\\rowcolor{gray!10}
G-Tuning~\cite{G-Tuning} & Dec 2023 & AAAI'24 & Fine-tuning & \trainicon Miscellaneous \tuneicon SFT & Motifs & Graph classification \\
SUPT~\cite{SUPT} & Feb 2024 & arXiv & Prompting & \trainicon Miscellaneous \tuneicon SFT & Node features; context & Graph classification \\\rowcolor{gray!10}
GraphPAR~\cite{GraphPAR} & Feb 2024 & WWW'24 & \makecell[l]{Fine-tuning\\(PEFT)} & \makecell[l]{\trainicon Miscellaneous\\\tuneicon Node instance discrimination; SFT} & Node features & Node classification \\
IGAP~\cite{IGAP} & Feb 2024 & WWW'24 & Prompting & \trainicon Miscellaneous \tuneicon SFT & \makecell[l]{Node features; \\spectrum} & \makecell[l]{Node classification; \\graph classification }\\\rowcolor{gray!10}
Bridge-Tune~\cite{Bridge-Tune} & Mar 2024 & AAAI'24 & Fine-tuning & \trainicon Miscellaneous \tuneicon SFT & -- & \makecell[l]{Node classification; \\link prediction }\\
STGP~\cite{STGP} & May 2024 & CIKM'24 & Prompting & \trainicon Masked feature prediction & Node features & Time series forcasting \\\rowcolor{gray!10}
DyGPrompt~\cite{DyGPrompt} & May 2024 & ICLR'25 & Prompting & \makecell[l]{\trainicon (Temporal) link prediction \tuneicon SFT} & Node features; links & \makecell[l]{(Temporal)\\node classification; \\link prediction }\\
P2TAG~\cite{P2TAG} & Jul 2024 & KDD'24 & Prompting & \trainicon Masked language modeling \tuneicon SFT & \makecell[l]{Node features; context; \\long-range similarities} & Node classification \\\rowcolor{gray!10}
TGPT~\cite{TGPT} & Aug 2024 & KDD'24 & Prompting & \trainicon Miscellaneous \tuneicon SFT & Motifs; global structure & \makecell[l]{Graph classification; \\graph regression }\\
GraphCLIP~\cite{GraphCLIP} & Oct 2024 & WWW'25 & Prompting & \makecell[l]{\trainicon Node-text discrimination \tuneicon SFT} & Node features & \makecell[l]{Node classification; \\link prediction }\\\rowcolor{gray!10}
HG-Adapter~\cite{HG-Adapter} & Nov 2024 & ICLR'25 & \makecell[l]{Fine-tuning\\(PEFT)} & \makecell[l]{\trainicon Miscellaneous\\\tuneicon Feature prediction; SFT} & Node features; context & \makecell[l]{(Heterogeneous)\\node classification; \\node clustering }\\
IA-GPL~\cite{IA-GPL} & Nov 2024 & TMLR'25 & Prompting & \trainicon Miscellaneous \tuneicon SFT & Node features & Graph classification \\\rowcolor{gray!10}
DAGPrompT~\cite{DAGPrompT} &Jan 2025 & WWW'25 & \makecell[l]{Fine-tuning\\(PEFT); \\prompting} & \makecell[l]{\trainicon Link prediction \tuneicon SFT} & Node features; context & \makecell[l]{Node classification; \\graph classification }\\
EdgePrompt~\cite{EdgePrompt} & Jan 2025 & ICLR'25 & Prompting & \trainicon Miscellaneous \tuneicon SFT & Links & \makecell[l]{Node classification; \\graph classification }\\\rowcolor{gray!10}
FPrompt~\cite{FPrompt} & Jan 2025 & WWW'25 & \makecell[l]{Fine-tuning\\(PEFT); \\prompting} & \makecell[l]{\trainicon Node instance discrimination \\\hspace{7.5pt} / node-graph discrimination \tuneicon SFT} & Node features; links & Node classification \\
\bottomrule[1.2pt]
\end{tabular}
}
\end{table*}

\begin{table*}[!t]
\setlength{\tabcolsep}{3pt}
\caption{
\conditionalblue{
Summary of graph language models.\\
``Adapter'' refers to all kinds of non-GNN/GT network modules apart from the LLM backbone, e.g., LoRA~\cite{LoRA} and linear projectors.\\
``AR'' and ``MLM'' refer to autoregressive and masked language modeling, respectively. ``SFT'' refers to ``supervised fine-tuning''. 
}
}\label{glmtable}
\centering
\adjustbox{max width=1.0\textwidth}{
\begin{tabular}{p{2.7cm}lp{2cm}p{4.6cm}p{2.6cm}p{4.6cm}p{2.2cm}p{3.5cm}}
\toprule[1.2pt]
\multirow{2.4}{*}{Model} & \multirow{2.4}{*}{Time} & \multirow{2.4}{*}{Venue} & \multicolumn{3}{c}{\trainicon Training/\tuneicon tuning tasks} & \multirow{2.4}{*}{\makecell[l]{Graph knowledge\\focused on}} & \multirow{2.4}{*}{Downstream tasks} \\[-1pt]
\arrayrulecolor[RGB]{180, 180, 180}\cmidrule(){4-6}\arrayrulecolor[RGB]{0, 0, 0} 
& & & LM/LLM & Adapter & GNN/GT & & 
\\[-1pt] \cmidrule(){1-8}
KEPLER~\cite{KEPLER} & Nov 2019 & TACL'21 & \makecell[l]{\trainicon MLM;\\\hspace{7.5pt} node instance discimination} & -- & -- & \makecell[l]{Node features;\\links} & Link prediction \\\rowcolor{gray!10}
GraphFormers~\cite{GraphFormers} & May 2021 & NeurIPS'21 & \trainicon Link prediction & -- & \trainicon Link prediction & Links & Link prediction \\
GIANT~\cite{GIANT} & Nov 2021 & ICLR'22 & \tuneicon Neighborhood prediction & -- & -- & \makecell[l]{Context;\\clusters} & Node classification \\\rowcolor{gray!10}
GLEM~\cite{GLEM} & Oct 2022 & ICLR'23 & \tuneicon SFT & -- & \tuneicon SFT & -- & Node classification \\
G2P2~\cite{G2P2} & May 2023 & SIGIR'23 & \makecell[l]{\trainicon Node-text discrimination\\\hspace{7.5pt} context discrimination \tuneicon SFT} & -- & \makecell[l]{\trainicon Node-text discrimination\\\hspace{7.5pt} context discrimination} & \makecell[l]{Node features;\\context} & Node classification \\\rowcolor{gray!10}
NLGraph~\cite{NLGraph} & May 2023 & NeurIPS'23 & Frozen & -- & -- & -- & Graph question answering \\
Patton~\cite{Patton} & May 2023 & ACL'23 & \makecell[l]{\trainicon MLM; link prediction} & -- & \makecell[l]{\trainicon MLM; link prediction} & \makecell[l]{Node features;\\links;\\context} & \makecell[l]{Node classification;\\link prediction; etc.} \\\rowcolor{gray!10}
ConGraT~\cite{ConGraT} & May 2023 & \makecell[l]{ACL\\Workshop\\(TextGraphs)'24} & \trainicon Node-text discrimination & -- & \trainicon Node-text discrimination & \makecell[l]{Node features;\\long-range\\similarities} & \makecell[l]{Node classification;\\link prediction} \\
TAPE~\cite{TAPE} & May 2023 & ICLR'24 & Frozen (LLM) \tuneicon SFT (LM) & -- & \tuneicon Miscellaneous & Node features & \makecell[l]{Node classification;\\link prediction} \\\rowcolor{gray!10}
GALM~\cite{GALM} & Jun 2023 & KDD'23 & \trainicon Link prediction \tuneicon SFT & -- & \trainicon Link prediction \tuneicon SFT & Links & \makecell[l]{(Heterogeneous)\\node classification;\\link prediction;\\edge classification} \\
KEA~\cite{KEA} & May 2023 & \makecell[l]{KDD\\Explorations\\Newsletter'24} & Frozen (LLM) \tuneicon SFT (LM) & -- & \tuneicon Miscellaneous & Node features & Node classification \\\rowcolor{gray!10}
LLM4Mol~\cite{LLM4Mol} & Jul 2023 & arXiv & Frozen (LLM) \tuneicon SFT (LM) & -- & -- & \makecell[l]{Motifs;\\global structure} & Graph classification \\
SimTeG~\cite{SimTeG} & Aug 2023 & arXiv & Frozen & \tuneicon SFT & \tuneicon SFT & -- & \makecell[l]{Node classification;\\link prediction} \\\rowcolor{gray!10}
InstructGLM~\cite{InstructGLM} & Aug 2023 & \makecell[l]{EACL\\Findings'24} & \tuneicon Link prediction; SFT & -- & -- & Links; context & Node classification \\
G-Prompt~\cite{G-Prompt} & Sept 2023 & arXiv & Frozen & -- & \tuneicon MLM & \makecell[l]{Node features;\\context} & Node classification \\\rowcolor{gray!10}
WalkLM~\cite{WalkLM} & Sept 2023 & NeurIPS'23 & \tuneicon MLM & -- & -- & \makecell[l]{Long-range\\similarities} & \makecell[l]{Node classification;\\link prediction} \\
GNP~\cite{GNP} & Sept 2023 & AAAI'24 & Frozen & \makecell[l]{\tuneicon Link prediction;\\SFT} & \tuneicon Link prediction; SFT & Links & Graph question answering \\\rowcolor{gray!10}
OFA~\cite{OFA} & Oct 2023 & ICLR'24 & Frozen & -- & \tuneicon SFT & \makecell[l]{Node features;\\context} & \makecell[l]{Node classification;\\graph classification;\\link prediction} \\
GraphText~\cite{GraphText} & Oct 2023 & arXiv & Frozen & -- & -- & \makecell[l]{Node features;\\context} & Node classification \\\rowcolor{gray!10}
LLM-GNN~\cite{LLM-GNN} & Oct 2023 & ICLR'24 & Frozen & -- & \tuneicon Label-free node classification & Clusters & \makecell[l]{Node classification;\\link prediction} \\
GraphLLM~\cite{GraphLLM} & Oct 2023 & arXiv & Frozen & \tuneicon SFT & \tuneicon SFT & Node features & Graph question answering \\\rowcolor{gray!10}
LLM4NG~\cite{LLM4NG} & Oct 2023 & AAAI'25 & Frozen & \tuneicon Link prediction & \tuneicon Miscellaneous & Links & Node classification \\
GraphGPT~\cite{GraphGPT} & Oct 2023 & SIGIR'24 & \makecell[l]{Frozen (LLM)\\\trainicon Node-text discrimination (LM)} & \makecell[l]{\tuneicon Graph-instruction\\\hspace{7.5pt} matching; SFT} & \trainicon Node-text discrimination & \makecell[l]{Node features;\\context} & \makecell[l]{Node classification;\\link prediction} \\\rowcolor{gray!10}
GRENADE~\cite{GRENADE} & Oct 2023 & \makecell[l]{EMNLP\\Findings'23} & \makecell[l]{\trainicon Node instance discrimination;\\\hspace{7.5pt} context discrimination} & -- & \makecell[l]{\trainicon Node instance discrimination;\\context discrimination} & \makecell[l]{Node features;\\context} & \makecell[l]{Node classification;\\node clustering;\\link prediction} \\
LLM4DyG~\cite{LLM4DyG} & Oct 2023 & KDD'24 & Frozen & -- & -- & -- & \makecell[l]{(Temporal)\\graph question answering} \\\rowcolor{gray!10}
DGTL~\cite{DGTL} & Oct 2023 & arXiv & Frozen & -- & \tuneicon SFT & Context & Node classification \\
THLM~\cite{THLM} & Nov 2023 & \makecell[l]{EMNLP\\Findings'23} & \makecell[l]{\trainicon MLM; context discrimination\\\tuneicon SFT} & -- & \makecell[l]{\trainicon MLM; context discrimination} & \makecell[l]{Node features;\\context} & \makecell[l]{(Heterogeneous)\\node classification;\\link prediction} \\\rowcolor{gray!10}
Sun {\it et al.}~\cite{A-D+LPA} & Nov 2023 & arXiv & Frozen & -- & \tuneicon SFT & Links & Node classification \\
LEADING~\cite{LEADING} & Dec 2023 & arXiv & \tuneicon Miscellaneous & -- & \tuneicon Miscellaneous & -- & Node classification \\\rowcolor{gray!10}
ENGINE~\cite{ENGINE} & Jan 2024 & IJCAI'24 & Frozen & -- & \tuneicon SFT & -- & \makecell[l]{Node classification;\\link prediction} \\
SNS~\cite{SNS} & Feb 2024 & arXiv & Frozen & -- & -- & \makecell[l]{Long-range\\similarities} & Node classification \\\rowcolor{gray!10}
GraphToken~\cite{GraphToken} & Feb 2024 & arXiv & Frozen & -- & \tuneicon SFT & Node features & Graph question answering \\
LinguGKD~\cite{LinguGKD} & Feb 2024 & AAAI'25 & Frozen & \tuneicon SFT & \makecell[l]{\tuneicon Node instance discrimination;\\\hspace{7.5pt} SFT} & \makecell[l]{Node features;\\node properties;\\context} & Node classification \\\rowcolor{gray!10}
GraphTranslator~\cite{GraphTranslator} & Feb 2024 & WWW'24 & Frozen & \makecell[l]{\tuneicon Node-text\\\hspace{7.5pt} discrimination;\\\hspace{7.5pt} masked language\\\hspace{7.5pt} modeling} & \trainicon Link prediction & \makecell[l]{Node features;\\context} & \makecell[l]{Node classification;\\graph question answering} \\
LLaGA~\cite{LLaGA} & Feb 2024 & ICML'24 & Frozen & \tuneicon SFT & -- & Context & \makecell[l]{Node classification;\\link prediction} \\\rowcolor{gray!10}
InstructGraph~\cite{InstructGraph} & Feb 2024 & \makecell[l]{ACL\\Findings'24} & Frozen & \tuneicon SFT & -- & Miscellaneous & \makecell[l]{Node classification;\\link prediction;\\graph question answering} \\
GraphPrompter~\cite{GraphPrompter} & Feb 2024 & \makecell[l]{WWW'24\\(short papers)} & Frozen & -- & \tuneicon SFT & Context & \makecell[l]{Node classification;\\link prediction} \\\rowcolor{gray!10}
Pan {\it et al.}~\cite{DistillLLM} & Feb 2024 & CIKM'24 & Frozen & -- & \makecell[l]{\tuneicon Node instance discrimination;\\\hspace{7.5pt} label-free node classification} & \makecell[l]{Node features;\\context;\\long-range\\similarities} & Node classification \\
GraphAdapter~\cite{GraphAdapter} & Feb 2024 & WWW'24 & Frozen & -- & \tuneicon AR; SFT & \makecell[l]{Node features;\\context} & Node classification \\\rowcolor{gray!10}
UniGraph~\cite{UniGraph} & Feb 2024 & arXiv & \makecell[l]{Frozen (LLM)\\\trainicon Node instance discrimination;\\\hspace{7.5pt} MLM (LM)} & \tuneicon SFT & \makecell[l]{\trainicon MLM;\\\hspace{7.5pt} node instance discrimination} & Node features & \makecell[l]{Node classification;\\graph classification;\\edge classification} \\
HiGPT~\cite{HiGPT} & Feb 2024 & KDD'24 & \makecell[l]{Frozen (LLM)\\\trainicon Node-text discrimination (LM)} & \makecell[l]{\tuneicon Graph-instruction\\\hspace{7.5pt} matching; SFT} & \trainicon Node-text discrimination & \makecell[l]{Node features;\\context} & \makecell[l]{(Heterogeneous)\\node classification} \\\rowcolor{gray!10}
GraphWiz~\cite{GraphWiz} & Feb 2024 & KDD'24 & \tuneicon SFT & -- & -- & Miscellaneous & Graph question answering \\
OpenGraph~\cite{OpenGraph} & Mar 2024 & \makecell[l]{EMNLP\\Findings'24} & Frozen & -- & \tuneicon Masked link prediction & \makecell[l]{Links;\\long-range\\similarities} & \makecell[l]{Node classification;\\link prediction} \\\rowcolor{gray!10}
GraphInstruct~\cite{GraphInstruct} & Mar 2024 & arXiv & Frozen & \tuneicon SFT & -- & Miscellaneous & Graph question answering \\
LOGIN~\cite{LOGIN} & May 2024 & WSDM'25 & Frozen & -- & \tuneicon SFT & \makecell[l]{Node features;\\links} & Node classification \\\rowcolor{gray!10}
TAGA~\cite{TAGA} & May 2024 & arXiv & Frozen & -- & \tuneicon Node-text discrimination & \makecell[l]{Node features;\\context} & Node classification \\
\bottomrule[1.2pt]
\end{tabular}
}
\end{table*}

\setcounter{table}{2}
\begin{table*}[!t]
\setlength{\tabcolsep}{3pt}
\caption{
\conditionalblue{
Summary of graph language models (continued).\\
``Adapter'' refers to all kinds of non-GNN/GT network modules apart from the LLM backbone, e.g., LoRA~\cite{LoRA} and linear projectors.\\
``AR'' and ``MLM'' refer to autoregressive and masked language modeling, respectively. ``SFT'' refers to ``supervised fine-tuning''. 
}
}
\centering
\adjustbox{max width=1.0\textwidth}{
\begin{tabular}{p{2.7cm}lp{2cm}p{4.6cm}p{2.6cm}p{4.6cm}p{2.2cm}p{3.5cm}}
\toprule[1.2pt]
\multirow{2.4}{*}{Model} & \multirow{2.4}{*}{Time} & \multirow{2.4}{*}{Venue} & \multicolumn{3}{c}{\trainicon Training/\tuneicon tuning tasks} & \multirow{2.4}{*}{\makecell[l]{Graph knowledge\\focused on}} & \multirow{2.4}{*}{Downstream tasks} \\[-1pt]
\arrayrulecolor[RGB]{180, 180, 180}\cmidrule(){4-6}\arrayrulecolor[RGB]{0, 0, 0} 
& & & LM/LLM & Adapter & GNN/GT & & 
\\[-1pt] \cmidrule(){1-8}
GAugLLM~\cite{GAugLLM} & Jun 2024 & KDD'24 & \makecell[l]{Frozen (LLM)\\\tuneicon Neighborhood prediction} & -- & \makecell[l]{\tuneicon Node instance discrimination\\\hspace{7.5pt} / masked feature prediction} & \makecell[l]{Node features;\\context} & Node classification \\
UniGLM~\cite{UniGLM} & Jun 2024 & WSDM'25 & \trainicon Node instance discrimination & -- & \tuneicon Miscellaneous & \makecell[l]{Node features;\\context} & \makecell[l]{Node classification;\\link prediction} \\\rowcolor{gray!10}
GSPT~\cite{GSPT} & Jun 2024 & LoG'24 & \trainicon Masked feature prediction & -- & -- & \makecell[l]{Node features;\\context} & \makecell[l]{Node classification;\\link prediction} \\
HIGHT~\cite{HIGHT} & Jun 2024 & arXiv & Frozen & \makecell[l]{\tuneicon Masked feature\\\hspace{7.5pt} prediction; SFT} & Frozen & \makecell[l]{Node features;\\motifs} & Graph classification; etc. \\\rowcolor{gray!10}
GOFA~\cite{GOFA} & Jul 2024 & ICLR'25 & Frozen & -- & \makecell[l]{\tuneicon AR; similarity prediction;\\\hspace{7.5pt} common neighbor prediction;\\\hspace{7.5pt} SFT} & \makecell[l]{Node features;\\context;\\long-range\\similarities} & \makecell[l]{Node classification;\\link prediction} \\
P2TAG~\cite{P2TAG} & Jul 2024 & KDD'24 & \trainicon MLM & -- & \trainicon MLM \tuneicon SFT & \makecell[l]{Node features;\\context;\\long-range\\similarities} & Node classification \\\rowcolor{gray!10}
Path-LLM~\cite{Path-LLM} & Aug 2024 & arXiv & Frozen & \trainicon AR & -- & \makecell[l]{Long-range\\similarities} & \makecell[l]{Node classification;\\link prediction} \\
LLM4RGNN~\cite{LLM4RGNN} & Aug 2024 & KDD'25 & Frozen & \tuneicon Link prediction & \tuneicon Miscellaneous & Links & Node classification \\\rowcolor{gray!10}
CMRP~\cite{CMRP} & Aug 2024 & KDD'24 & Frozen & \tuneicon SFT & \tuneicon SFT & Links & \makecell[l]{Node classification;\\link prediction;\\graph classification;\\graph question answering} \\
TEA-GLM~\cite{TEA-GLM} & Aug 2024 & NeurIPS'24 & Frozen & \tuneicon SFT & \makecell[l]{\tuneicon Node instance discrimination;\\dimension discrimination} & Node features & \makecell[l]{Node classification;\\link prediction} \\\rowcolor{gray!10}
Skianis {\it et al.}~\cite{Pseudo-Code} & Sept 2024 & arXiv & Frozen & -- & -- & -- & Graph question answering \\
GUNDAM~\cite{GUNDAM} & Sept 2024 & arXiv & \tuneicon Similarity prediction; SFT & -- & -- & \makecell[l]{Long-range\\similarities} & Graph question answering \\\rowcolor{gray!10}
AuGLM~\cite{AuGLM} & Oct 2024 & arXiv & \tuneicon SFT & -- & -- & \makecell[l]{Context;\\long-range\\similarities} & Node classification \\
AskGNN~\cite{AskGNN} & Oct 2024 & \makecell[l]{EMNLP\\Findings'24} & Frozen & -- & \makecell[l]{\tuneicon Node instance discrimination;\\\hspace{7.5pt} SFT} & \makecell[l]{Node features;\\long-range\\similarities} & Node classification \\\rowcolor{gray!10}
GraphCLIP~\cite{GraphCLIP} & Oct 2024 & WWW'25 & Frozen & -- & \tuneicon Node-text discrimination; SFT & Node features & \makecell[l]{Node classification;\\link prediction} \\
LLaSA~\cite{LLaSA} & Nov 2024 & arXiv & \makecell[l]{\trainicon AR;\\node-text discrimination} & \tuneicon SFT & \makecell[l]{\trainicon AR;\\node-text discrimination} & Node features & Graph question answering \\\rowcolor{gray!10}
TANS~\cite{TANS} & Dec 2024 & NAACL'25 & Frozen & -- & \tuneicon Miscellaneous & \makecell[l]{Node properties;\\context} & Node classification \\
Locle~\cite{Locle} & Dec 2024 & SIGIR'25 & Frozen & -- & \tuneicon Label-free node classification & Links; clusters & Node classification \\\rowcolor{gray!10}
HierPromptLM~\cite{HierPromptLM} & Jan 2025 & arXiv & \makecell[l]{\tuneicon MLM;\\link prediction} & -- & -- & \makecell[l]{Node features;\\links} & \makecell[l]{(Heterogeneous)\\node classification;\\link prediction} \\
SFGL~\cite{SFGL} & Jan 2025 & ICLR'25 & Frozen (LLM) \tuneicon SFT (LM) & -- & \tuneicon Miscellaneous & \makecell[l]{Long-range\\similarities} & Node classification \\
\bottomrule[1.2pt]
\end{tabular}
}
\end{table*}

\end{document}